# Can LLM-Simulated Respondents Pre-Test Survey Questions? Evidence From Preregistered Prospective Validation

Sristi Bafna, Mac MacComas, Michael Bader, Kristina Gligorić
Johns Hopkins University

**Abstract**

Large language models (LLMs) can be conditioned to simulate behavioral responses, including survey responses, but it remains unclear whether and how these simulations can support downstream research tasks. Here, we evaluate whether simulated respondents can support pre-testing new candidate survey questions before fielding them, operationalized as prioritizing candidate survey items. Specifically, we rank candidate questions by their mean responses, which can reveal ceiling or floor effects; their variance, which indicates whether responses contain meaningful variation; and differences between subpopulations, which indicate whether questions capture distinctions between relevant groups within a population. We introduce a preregistered, prospective evaluation that commits model predictions before target human responses exist, ruling out memorization of those responses by design. Using two waves of a local survey (2025: N = 1,271; 2026: N = 1,161), we built simulated respondents from 2025 participants and preregistered GPT-4.1 predictions for 22 questions introduced in 2026. Simulations accurately ranked questions by mean responses (Spearman $\rho = 0.71$) and anticipated which questions distinguished city from county residents ($\rho = 0.88$), Black from White residents ($\rho = 0.68$), and employed from unemployed residents ($\rho = 0.57$), suggesting simulations can help prioritize questions before fielding. More broadly, our prospective, preregistered design provides a general framework for evaluating AI simulations without contamination from memorized target outcomes.

## 1. Introduction

Large language models (LLMs) can be prompted to answer survey questions as if they were study participants. Growing literature argues that such simulations could change how social science is conducted (Bail, 2024; Anthis et al., 2025; Xu et al., 2026). Conditioned on a demographic profile, or on answers a person has given before, simulated respondents have reproduced the distribution of political attitudes and vote choice across subgroups (Argyle et al., 2023), replicated findings from behavioral economics, psychology, and management (Aher et al., 2023; Horton, 2023; Cui et al., 2025), and predicted the direction and relative size of treatment effects in survey experiments (Ashokkumar et al., 2026). When built from long interviews, they have approached the accuracy with which people reproduce their own survey answers two weeks later (Park et al., 2024). Similarly, models fine-tuned on survey data predicted distributions of public opinion (Suh et al., 2025). These results raise the possibility of applying AI simulations to augment or simulate survey-based research.

At the same time, simulations face several key limitations. First, when a model reproduces survey results, the AI may be modeling opinions and behaviors or it may be recalling results it has already seen in training (Ludwig et al., 2025). This distinction is critical because apparent agreement with human responses may overstate the model's ability to generalize to new survey questions. Evaluations based on previously published datasets therefore cannot cleanly separate inference from memorization, making prospective tests especially important. Additionally, simulated samples do not represent everyone equally well (Santurkar et al., 2023; Abeliuk et al., 2026; Hu et al., 2026), and they can stereotype and flatten the responses from groups they are asked to represent (Wang et al., 2025). AI simulations can also exaggerate such that simulated personas drift toward caricatures (Cheng et al., 2023; Ma et al., 2026), simulated answers vary less than humans do (Bisbee et al., 2024; Xie et al., 2026), and predicted effects are systematically over-estimated (Ashokkumar et al., 2026). Overall, the conditions under which simulated data may count as evidence remain contested (Hullman et al., 2026). These limitations may be substantial enough to mitigate against using simulated samples to support survey-based research at all (Pew Research Center, 2026; Rothschild et al., 2026).

An alternative to considering how well AI simulations globally represent populations as respondent replacements would be to consider which survey research tasks AI simulations can support. This more cautious approach treats AI simulations as imperfect black boxes rather than as a replacement for respondents (Hazra et al., 2026). In particular, AI simulations may be especially cost effective for helping researchers decide which questions to field (i.e., instrument design), while data collected from humans remain the basis for estimates and statistical inference (Egami et al., 2023; Broska et al., 2025; Rothschild et al., 2025; Anthis et al., 2025; Romberg et al., 2026). This approach corresponds to tools for testing instruments on virtual populations (Yu et al., 2026), like detecting flaws in the wording of draft questions (Olivos & Liu, 2024; Sturgis et al., 2026). Essentially, AI simulations may be good at replacing the expensive process of pre-testing questions and instruments through human-to-human cognitive testing that imposes large cognitive burdens on respondents (Schaeffer and Presser, 2003).

However, it is unclear whether AI simulations can support pre-testing. As mentioned above, evaluations of simulated answers mostly rely on questions that had already been asked, and often published, when the models were trained. The exceptions restrict evaluation to experimental studies that were not yet published, or to answers to the same questions given after a model's training cutoff (Ashokkumar et al., 2026; Jia et al., 2026). Overall, we lack prospective evidence that simulations can evaluate candidate survey questions. To address this gap we propose two innovations to support survey pretesting: we **1) operationalize pre-testing as a candidate ranking task** and **2) develop a novel pre-registration-based evaluation protocol**. We demonstrate the utility of these innovations with a preregistered, prospective study (2025 wave: N = 1,271; 2026 wave: N = 1,161).

First, we formulate pre-testing as a ranking problem. We distinguish between two components of survey pre-testing: *cognitive pre-testing*, which evaluates how respondents understand and answer individual questions (McColl, 2006), and *instrument-level pre-testing*, which helps researchers choose among a battery of candidate questions and design the instrument as a whole. We focus on the latter and formulate it as a ranking problem. Presser et al. (2004) characterize pre-testing as a process for evaluating questionnaires in advance and emphasize its role in identifying problems and improving the questionnaire before fielding. This makes pre-testing inherently consequential for instrument design: researchers use evidence collected during pre-testing to decide which candidate questions to retain, revise, or exclude. Because only so many questions may be included in a survey, questionnaire space is a scarce resource. Therefore, when researchers consider a battery of candidate questions, a central instrument-design decision is determining which questions are most useful—for example, which have sufficient variation, identify differences between respondents, or separate one sub-population from another. A researcher choosing among candidate questions therefore needs instrument pre-testing to reveal their relative usefulness, rather than merely reproduce exact response percentages.

We therefore formulate this downstream decision as a ranking problem. We focus on three properties that inform whether a candidate question is likely to be useful in an instrument. Ranking questions by their **mean responses** can identify questions whose answers cluster near the ceiling or floor of the response scale; ranking by **variance** can identify questions that elicit little information because respondents answer nearly uniformly; and ranking by **differences between** **subpopulations** can identify questions that are most sensitive to meaningful differences between groups within the target population. AI simulations present a potential cost-effective process for this ordering task, but must be evaluated directly on this downstream objective.

Second, we introduce a preregistration-based prospective evaluation. Our design takes advantage of a specific moment in the survey lifecycle: after the instrument has been finalized but before data collection begins, when the target human responses do not yet exist. This creates a natural opportunity for prospective evaluation, and preregistration provides the commitment mechanism needed to use it. We generate simulated responses during this window and preregister the model, prompt, predictions, and analysis before the survey is fielded. Human responses are collected only afterward. Because the target outcomes did not exist when the predictions were committed, agreement between simulated and human responses cannot be explained by memorization of those outcomes. At the same time, preregistration constrains analytical flexibility by fixing key modeling and evaluation choices in advance and provides greater transparency into the research process (Nosek et al., 2018; Kosch & Feger, 2026; Vaccaro, 2026; Gelman et al., 2013).

Using the Baltimore Area Survey, an annual, repeated cross-sectional survey of representative of residents of Baltimore City and Baltimore County (two independent municipal jurisdictions in Maryland), we constructed a synthetic sample with one simulated respondent for each human respondent to the 2025 wave (N = 1,271), conditioned on that respondent's demographic profile and their answers to 258 questions, and prompted GPT-4.1 to answer the questions that were new in 2026. Before the 2026 wave was fielded, we pre-registered three predictions for each question new in the 2026 wave (referred to as *candidate questions*), based on the simulated responses from the synthetic sample. Specifically, we pre-registered each question's mean answer, the variance of its answers, and the difference in answers between groups along five demographic axes (jurisdiction, race, political ideology, employment, and ability to afford transportation). After the 2026 wave was fielded (N = 1,161), we test whether ranking questions based on AI-generated responses produces the same ordering as ranking questions based on real human responses, using rank correlations with thresholds fixed in advance. As a secondary analysis, we ran the same tests on two fine-tuned versions of the model.

We find that simulated respondents can rank candidate questions on a number of properties, with important limitations. Simulated respondents ranked the questions by mean answer (Spearman ρ

= 0.71), anticipated which questions separate city from county residents (ρ = 0.88), Black from White residents (ρ = 0.68), and full-time employed from unemployed residents (ρ = 0.57), and in which direction. Question responses from simulated respondents were not ranked by how much answers vary (ρ = 0.31) nor by differences between conservatives and liberals (ρ = 0.08); fine-tuning, however, improved the variance ranking. The rank-order agreement was positively correlated despite the synthetic sample having point estimates that were statistically distinguishable from the human sample.

Overall, our results open the possibility of using AI simulations as a screening step in early stages of survey development. Before a question is fielded, a simulation can indicate where it is likely to fall relative to other candidates, and whether it is likely to separate groups defined by place, race, or employment, so that limited pre-testing effort and limited space in the questionnaire go to the most promising candidate questions. Following our study design and registering predictions before a survey is fielded offers a way to evaluate AI simulations that is free of memorization, and it can be repeated in any survey that adds new questions.

## 2. Results

### 2.1 Study design overview

The BAS is an annual survey taken of repeated cross-sections representative of adults living in Baltimore City and Baltimore County, fielded by the 21st Century Cities Initiative. We preregistered the study on July 8, 2026, before the 2026 wave of the Baltimore Area Survey (BAS) was fielded. The preregistration fixed the simulated responses, the predicted value of every quantity for every candidate question, the hypotheses, the tests, and the decision thresholds.

We built one simulated respondent for each of 1,271 respondents to the 2025 wave. Each prompt contained that respondent's rich demographic profile and their 2025 answers [258 questions]. Responses were generated by GPT-4.1 (gpt-4.1-2025-04-14) without fine-tuning, at temperature 0.3, with one prompt per simulated respondent. Simulated respondents were then prompted to answer the candidate 2026 questions with their answer options.

The preregistration designated 22 questions introduced in the 2026 wave as the confirmatory set. The questions are either completely new, use new scales, or new in the 2026 wave compared to the previous wave. The unit of analysis is a question. For each question, we registered three quantities calculated based on simulated respondents: the mean answer on a 0–1 scale (H1a; i.e., predicting average response to each question), the variance of answers on the same scale (H1b; i.e., predicting scale-normalized variance), and the signed difference between two groups (H1c; i.e., predicting if one of the two groups will give a higher answer, and by how much). The group difference is measured as the coefficient β from a survey-weighted bivariate regression of the standardized answer on a binary group indicator. We include multiple axes: jurisdiction (city vs. county), race (Black vs. White), ideology (conservative vs. liberal), and employment (full-time vs. unemployed).

For example, for a candidate question asking how satisfied respondents are with businesses in or near their neighborhood (five points, from "not at all" to "extremely satisfied"), on a scale 0-1, the simulated respondents predicted a mean of 0.55 (between "somewhat" and "very satisfied"), a variance of 0.03, and city residents answering 0.65 standard deviations lower than county residents ($\beta = -0.65$).

The 2026 wave successfully collected responses from 1,161 residents. The data analysis was conducted following the pre-registered data analysis strategy using 2026 survey weights. All questions met the inclusion threshold.

### 2.2 Preregistered analyses

| **Hypothesis** | **Zero-shot** | **Fine-tuned on 2025** | **Fine-tuned on 2024** |
|---|---|---|---|
| H1a: mean answer | Supported: $\rho = 0.71$, $p < 0.001$ | Supported: $\rho = 0.69$, $p < 0.001$ | Supported: $\rho = 0.64$, $p < 0.001$ |
| H1b: variance | Not supported: $\rho = 0.31$, $p = 0.079$ | Supported: $\rho = 0.44$, $p = 0.019$ | Supported: $\rho = 0.43$, $p = 0.023$ |
| H1c-1: city vs. county | Supported: $\rho = 0.88$, $p < 0.001$ | Supported: $\rho = 0.89$, $p < 0.001$ | Supported: $\rho = 0.75$, $p < 0.001$ |
| H1c-2: Black vs. White | Supported: $\rho = 0.68$, $p < 0.001$ | Supported: $\rho = 0.71$, $p < 0.001$ | Supported: $\rho = 0.66$, $p < 0.001$ |
| H1c-3: conservative vs. liberal | Not supported: $\rho = 0.08$, $p = 0.358$ | Not supported: $\rho = 0.11$, $p = 0.313$ | Supported: $\rho = 0.54$, $p = 0.005$ |
| H1c-4: full-time vs. unemployed | Supported: $\rho = 0.57$, $p = 0.003$ | Supported: $\rho = 0.57$, $p = 0.003$ | Supported: $\rho = 0.55$, $p = 0.004$ |

**Table 1. Confirmatory tests on the 22 candidate questions.** Green: supported; pink: not supported. H1a and H1b are tested at $\alpha = 0.05$ and H1c at $\alpha = 0.01$ (one-sided). The zero-shot model is the confirmatory test; the fine-tuned models are secondary.

### *Primary confirmatory hypotheses*

Table 1 summarizes the confirmatory tests. For the zero-shot model, four of the six are supported; for the two fine-tuned models, five and six are. Appendix A1 lists every prediction and observation.

**H1a: question means.** First, we find that simulated respondents rank candidate questions by their mean answer ($\rho = 0.71$, $p < 0.001$, $n = 22$). H1a is supported---a researcher comparing candidate questions would have known, before fielding, which would be answered high and which low. Predicted means are also close for most questions [median absolute error = 0.06 on the 0–1 scale; 14 of 22 questions within 0.10] (Figure 1; Figure 2, left). The largest gaps are on walking frequency, where residents walk less often than predicted (normalized mean: 0.30 predicted vs. 0.67 observed), on where residents would prefer to live in three years (0.20 vs. 0.51), and on whether windows were opened in the past 24 hours (64% predicted vs. 37.5% observed). The result does not depend on any single question (95% bootstrap interval: 0.34 to 0.90; Appendix A12), and two statistical baselines built from 2025 responses alone do not come close ($\rho = 0.22$ and 0.19; Appendix A14).

Agreement in ranking among candidate questions does not imply agreement in means for each question. Simulated participants differed on specific individual questions, and the differences are informative. For instance, simulated respondents expected finding an equivalent home to be harder than respondents reported (below the diagonal, Figure 2; left): the predicted mean of 0.59 falls between "somewhat" and "very difficult," whereas the observed mean of 0.40 falls between "a little" and "somewhat difficult." They also overestimated how often respondents walk: the predicted mean of 0.30 corresponds to walking a few times a week, whereas the observed mean of 0.67 corresponds to a few times in the past month (above the diagonal, Figure 2; left).

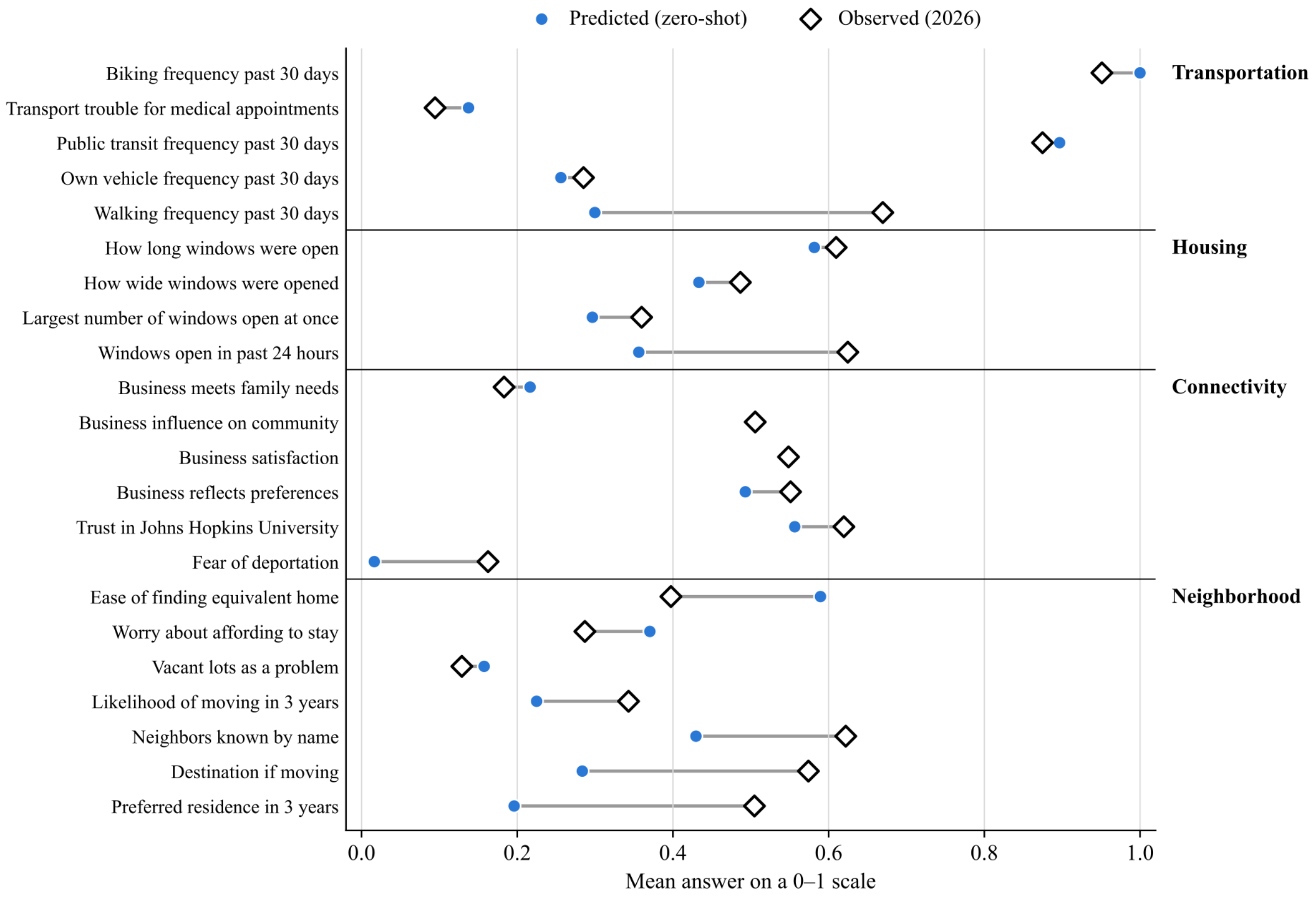

**Figure 1. Zero-shot predicted and observed mean answers for the 22 questions, by module.** Means are on a 0–1 scale. Within each module, questions are sorted by the difference between the observed and the predicted mean.

**H1b: question variances.** The zero-shot model ranks questions by variance in the predicted direction, but not at the pre-registered threshold ($\rho = 0.31$, $p = 0.079$, $n = 22$; Figure 2, right). H1b is therefore not supported for the zero-shot model. Simulated respondents agree with each other more than people do (mean normalized variance: 0.06 predicted vs. 0.10 observed). Such reduced variance in simulated samples has been documented in AI simulations (Bisbee et al., 2024; Wang et al., 2025; Xie et al., 2026). Importantly, we find that fine-tuning on an earlier wave recovers the variance ranking, while still underestimating the level of variance (H2).

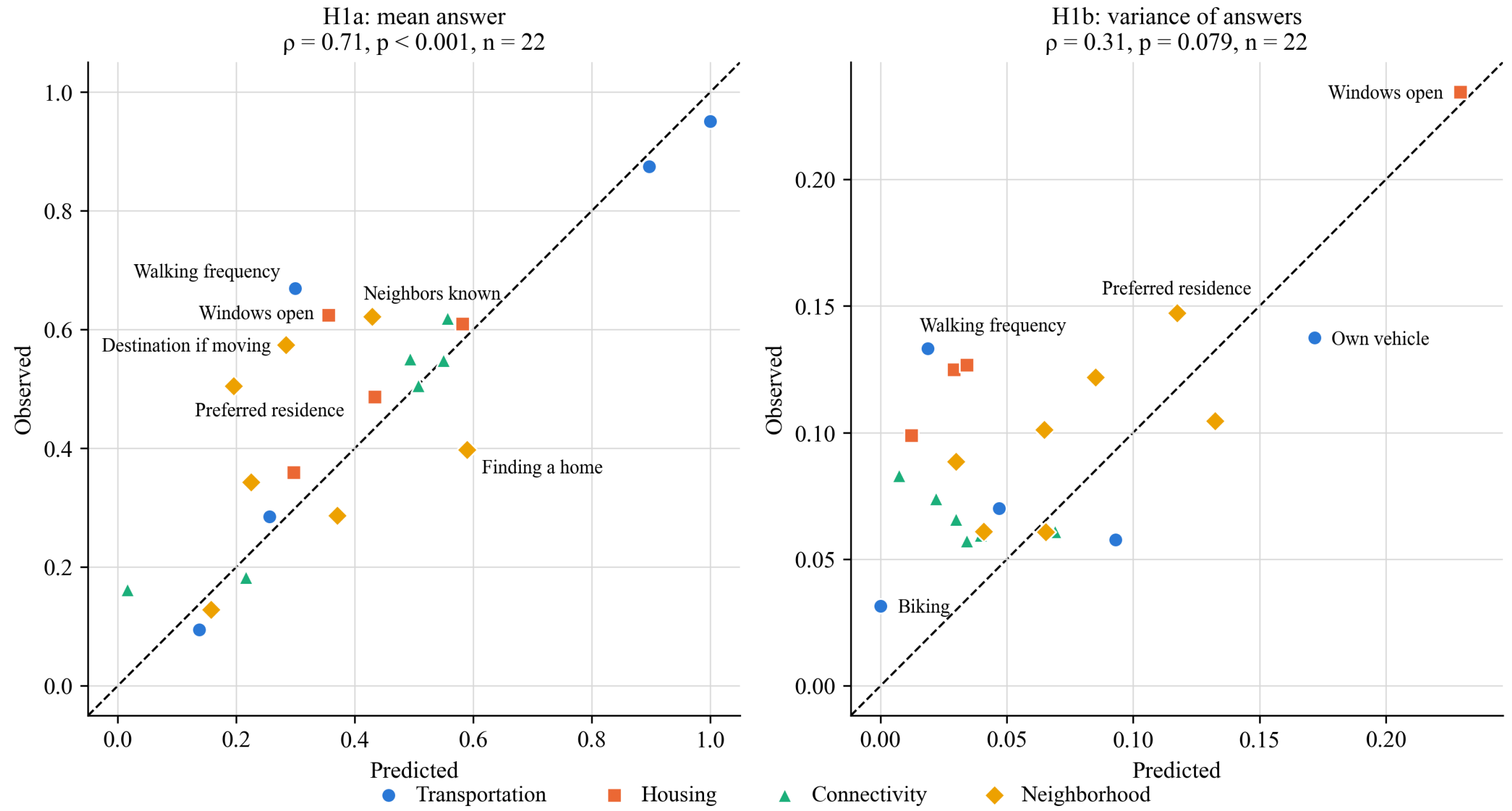

**Figure 2. Predicted and observed means (left, H1a) and variances (right, H1b) for the 22 questions, zero-shot model.** Each point corresponds to a candidate question. Points on the dashed diagonal are predicted exactly, and points far from the diagonal are not predicted correctly.

**H1c: group differences.** Simulated respondents anticipate which candidate questions separate groups, and in which direction, for three of the four tested axes (Figure 3). Jurisdiction ($\rho = 0.88$, $p < 0.001$), race ($\rho = 0.68$, $p < 0.001$), and employment ($\rho = 0.57$, $p = 0.003$) are supported at the corrected threshold ($\alpha = 0.01$), while ideology is not ($\rho = 0.08$, $p = 0.36$). Again, questions with largest differences are informative. For instance, for how long windows were kept open, simulated respondents predicted almost no city–county difference ($\beta = 0.07$), whereas city residents in fact kept them open longer ($\beta = 0.41$); they predicted that Black respondents would keep windows open for a shorter time than White respondents ($\beta = -0.41$), whereas the observed difference was small and in the opposite direction ($\beta = 0.11$), with one possible mechanism being access to air conditioning.

The ideology result matches our pilot on the 2024 wave, where ideology was also the only axis that failed (direction agreement observed on only 12 of 32 questions), and it matches reports that simulated samples represent some political groups less well than others (Santurkar et al., 2023; Abeliuk et al., 2026; Hu et al., 2026; Pew Research Center, 2026). Most of the candidate questions concern local, everyday matters such as businesses, windows, and travel, on which conservatives and liberals differ little (mean observed $|\beta| = 0.15$ SD). One interpretation is that on such questions, political orientation carries little information.

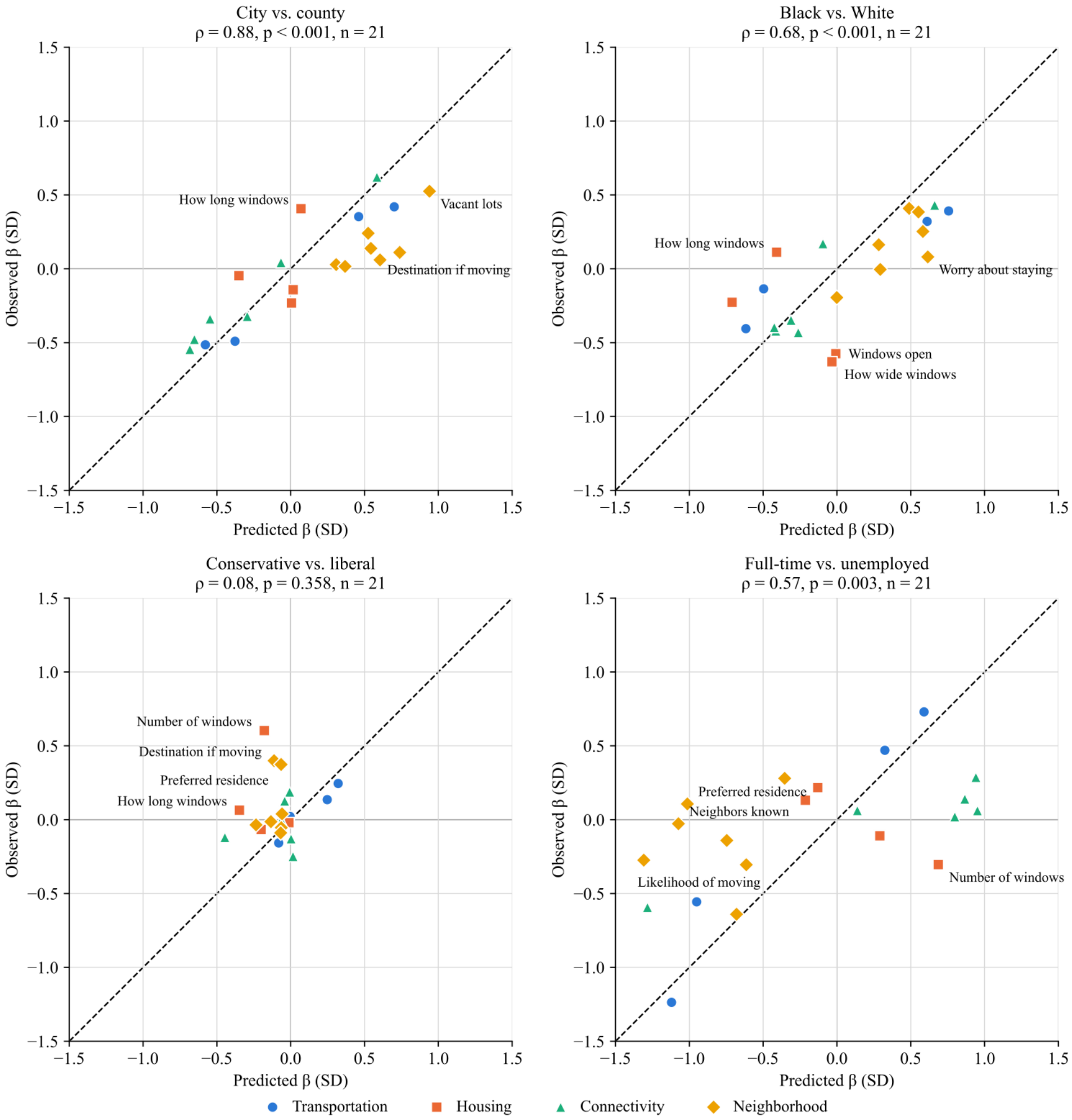

**Figure 3. Predicted and observed signed group differences (in standard deviations) for the 21 questions tested in H1c, zero-shot model.** Each point corresponds to a candidate question. Points on the dashed diagonal are predicted exactly, and points far from the diagonal are not predicted correctly.

**Comparison with the human split-half.** How high could these rank correlations be? The observed quantities are estimates from a sample, so even a perfect prediction would not reach $\rho = 1$. To establish an upper limit, we split the 2026 respondents at random into two halves, recomputed every observed quantity in each half, and correlated the two halves across questions (1,000 splits; Table 2). The human rankings of means and variances are almost free of noise (split-half $\rho = 0.98$ and 0.92), so the distance between the AI predictions and a perfect ranking on these quantities is AI model prediction error.

For jurisdiction (city vs county residents), the model is close to the human ceiling: it predicts one half of the respondents (ρ = 0.82) almost as well as the other half does (ρ = 0.88). For employment, two halves of the sample agree only moderately (ρ = 0.55), and the model predicts a half-sample (ρ = 0.51) about as well as the other half-sample does. For ideology (conservative vs. liberal), the human ranking is itself unreliable (ρ = 0.35): conservatives and liberals differ so little on these questions that their ordering is mostly noise. The zero-shot model still falls short of the ceiling on this axis (ρ = 0.08 vs. 0.35). Race is moderately predicted (ρ = 0.68; estimated ceiling 0.87). Details about the method and the fine-tuned models are listed in Appendix A13.

| Quantity | Human split-half ρ (ceiling) | Zero-shot ρ | Zero-shot vs. one half |
|---|---|---|---|
| Mean answer | 0.98 (0.97 to 0.99) | 0.71 | 0.71 |
| Variance | 0.92 (0.86 to 0.96) | 0.31 | 0.30 |
| City vs. county | 0.88 (0.78 to 0.95) | 0.88 | 0.82 |
| Black vs. White | 0.87 (0.75 to 0.94) | 0.68 | 0.68 |
| Conservative vs. liberal | 0.35 (0.05 to 0.60) | 0.08 | 0.03 |
| Full-time vs. unemployed | 0.55 (0.30 to 0.77) | 0.57 | 0.51 |

**Table 2. Reliability of the human rankings within data splits and comparison with the zero-shot model, on the 22 questions.** Split-half ρ captures the median Spearman correlation between two random halves of the 2026 respondents over 1,000 splits, with the range covering 95% of splits. Zero-shot vs. one half captures the mean correlation of the model's predictions with a half-sample. Low correlation between two random-halves of human respondents on political ideology (ρ = 0.35) provides additional contextualization for the failure to rank (ρ = 0.08). Cells are colored green for the hypothesis about correlation being supported, and pink for hypothesis not supported.

### *Secondary hypotheses*

**H2: fine-tuning.** We find that zero-shot prompting is as accurate as fine-tuning for most quantities. The models fine-tuned on 2025 and on 2024 data rank means no better than the zero-shot model (ρ = 0.69 and 0.64 vs. 0.71), and no better on jurisdiction (0.89 and 0.75 vs. 0.88), race (0.71 and 0.66 vs. 0.68), or employment (0.57 and 0.55 vs. 0.57). Fine-tuning improves the variance ranking, which passes the registered threshold for both fine-tuned models (ρ = 0.44, $p = 0.019$; ρ = 0.43, $p = 0.023$). On ideology the two fine-tuned models disagree (ρ = 0.11, $p = 0.31$; ρ = 0.54, $p = 0.005$).

For pre-testing, the cheaper approach, which needs no individual-level training data, is sufficient for ranking means and group differences. The gain on variance fits earlier evidence that fine-tuning on survey data improves predicted answer distributions (Suh et al., 2025). Appendix A3 compares the three models.

**H3: demographic axes.** Accuracy differs across axes in the order: jurisdiction, race, employment, ideology (ρ = 0.88, 0.68, 0.57, 0.08; direction agreement on 86%, 86%, 71%, and 52% of 21 questions). We expected identity axes (race and ideology) to be distorted by stereotyped portrayals. This holds for ideology and not for race, which is predicted well.

**H4: question types.** Accuracy differs across modules (mean absolute error of the predicted mean: Connectivity 0.05, Transportation 0.10, Housing 0.10, Neighborhood 0.17). Questions that can be answered from general patterns, or from related information about the respondent, are predicted closely. The largest errors are on questions whose answers depend on specific local conditions: the respondent's street (neighbors known by name: 0.43 predicted vs. 0.62 observed), the local housing market (ease of finding an equivalent home: 0.59 vs. 0.40), and the respondent's own plans to move (destination if moving: 0.28 vs. 0.57). Embodied questions are not uniformly harder: travel behavior other than walking is predicted well, while whether windows were opened on a given day is not. Appendix A9 gives the breakdown by module.

In summary, four of the six primary confirmatory hypotheses were supported for the zero-shot model, and five and six were supported for the two fine-tuned models. Before fielding, simulated respondents correctly anticipated which candidate questions would be answered high or low, and which would separate residents by jurisdiction, race and employment, with accuracy for jurisdiction and employment close to the agreement between two halves of the human sample.

### *Exploratory analyses*

**Robustness to the extent of novelty of the 22 candidate questions.** The candidate questions were new in 2026, but they were not all new in the same way. Comparing them with earlier waves puts them in three groups. Nine questions are **completely novel**: neither the question nor its response scale was used in any earlier wave. These are the **four questions about neighborhood businesses** (meeting family needs, influence on the sense of community, reflecting residents' preferences, and overall satisfaction), **fear of deportation**, **trust in Johns Hopkins University**, and **three follow-ups on window opening** (how wide, how long, and how many windows were open). Three questions are **partially novel**: the question appeared in past waves but with a different response scale. These are ease of finding an equivalent home, vacant lots as a problem, and transport trouble for medical appointments. The remaining ten questions are **novel relative to the previous wave**: they were not in the 2025 survey that the simulated respondents were conditioned on, but they were asked in 2024, two waves ago. They cover moving plans and preferences, knowing neighbors by name, worry about affording to stay, how often people walk, bike, take transit or drive, and whether any windows were open.

We repeated the confirmatory tests excluding questions that are novel relative to the previous wave ($n = 12$). The results were similar to the main analysis, and for most tests slightly stronger. The rank correlation between predicted and observed means rose from $\rho = 0.71$ to 0.80 ($p < 0.001$). The jurisdiction difference rose from 0.88 to 0.92 ($p < 0.001$) and the employment difference from 0.57 to 0.75 ($p = 0.003$), and all three remained significant. The race difference weakened from 0.68 to 0.58 ($p = 0.024$). As in the main analysis, the variance ranking and ideology differences were not supported. The fine-tuned models showed the same pattern. The questions that are completely new in the wave were not easier to predict than the others. Potential memorization of earlier waves therefore does not explain the model's performance.

**Full simulated distributions diverge, but rankings are informative.** A two-sample Kolmogorov–Smirnov test rejects equality of the simulated and observed distributions for most questions (not rejected for 2 questions with the zero-shot model, and for one and none with the fine-tuned models; $\alpha = 0.05$). Where distributions differ, simulated answers concentrate on one or two options while people spread across the scale (walking “a few times a week”: 85% simulated vs. 16% observed; “stay in your current home”: 75% vs. 30%). This is the same homogeneity that limits H1b, and it mirrors the gap in levels reported by Pew Research Center (2026). It supports evaluating simulations on ranking (the relevant objective for pre-testing), and not on point estimates. Appendix A4 shows the distributions.

**Temporal drift alone cannot explain simulation errors.** On repeated questions, we find that human answers are stable from year to year. Among 44 questions asked in both waves, the 2025 ranking almost reproduces the 2026 ranking [means: $\rho = 0.98$; variances: $\rho = 0.89$]. Drift between waves therefore does not explain the remaining prediction error on new questions. Appendix A8 plots the two waves.

**Simulated participants know too much.** The registered ordering of 11 neighborhoods by desirability matches the neighborhoods residents would seriously consider ($\rho = 0.88$) and would never consider ($\rho = 0.85$). Predicted shares order the neighborhoods well for "seriously consider" ($\rho = 0.90$ zero-shot; 0.86 and 0.79 fine-tuned), moderately for "never consider" (0.51; 0.33 and 0.74), and poorly for "do not know anything about" (−0.36; 0.20 and 0.34). Simulated respondents know too much: the zero-shot model predicts that 4% of residents know nothing about Homeland, against 56% observed.

This pattern may reflect a broader difficulty of LLMs in representing epistemic uncertainty: models tend to be overconfident and may prefer producing a substantive response over expressing uncertainty or abstaining when knowledge is limited (Kalai et al, 2026). Pew Research Center (2026) reports the same pattern on factual knowledge questions. The order of heating types in the home is also predicted well ($\rho = 0.82$ zero-shot; 0.86 and 0.96 fine-tuned; 7 types). We note these analyses are descriptive since they are based on $n = 11$ neighborhoods only (Appendix A7).

**Attitudes versus behaviors.** Under our classification, behaviors and experiences are ranked better than attitudes and perceptions ($\rho = 0.85$, $n = 9$ vs. $\rho = 0.63$, $n = 13$). See Appendix A9.

**Signed versus absolute differences.** Simulations predict the direction of group differences better than their size. Ranking questions by the absolute difference is significant for jurisdiction only ($|\beta|$: $\rho = 0.48$, $p = 0.014$; race: −0.05; ideology: 0.12; employment: 0.07; $n = 21$). This agrees with H3: the sign and the order are informative, but the magnitude is not (Appendix A10).

## 3. Discussion

We find that simulated respondents can rank survey questions that did not exist when the model was trained. The zero-shot model ranked 22 candidate questions by their mean answer [$\rho = 0.71$] and by the differences they elicit between city and county residents, Black and White residents, and employed and unemployed residents [$\rho = 0.88, 0.68, 0.57$]. Because the predictions were pre-registered before any response was collected, memorization cannot explain the result.

Interestingly, fine-tuning is mostly unnecessary. The zero-shot model matched the two fine-tuned models on means and on the jurisdiction, race, and employment differences. It fell behind on the variance ranking [$\rho = 0.31$ vs. 0.44 and 0.43], in line with evidence that fine-tuning on survey data improves predicted distributions (Suh et al., 2025). Pre-testing with simulations therefore does not require additional model training or holding-out data. A prompt that contains a simulated respondent's earlier simulated answers is enough, which makes the approach accessible and cost-efficient.

A questionnaire designer could use simulations as a screening step. One possible workflow could be to run the candidate battery through simulated respondents, shortlist questions by their expected mean and by the group differences they are expected to show, and spend cognitive interviews and pilot budget on the shortlist. This complements the use of LLMs to detect flawed wording (Sturgis et al., 2026) and fits professional guidance that keeps people involved in every AI-assisted step (Rothschild et al., 2026). The simulation does not replace the pilot, and it does not replace data collection: means, variances, and the size of group differences still have to come from human respondents.

Ranking is what instrument design needs, and it reconciles our result with more pessimistic ones. Studies that conclude simulated samples cannot stand in for respondents evaluated means and distributions (Bisbee et al., 2024; Perez et al., 2026; Pew Research Center, 2026; Xie et al., 2026). In this study, the same simulations that ranked questions well produced answer distributions that differ from the human ones for 20 of 22 questions, implying that correct ranking for pre-testing does not require strict distributional matches.

Preregistered prediction is a generalizable method. Any survey that adds new questions creates a natural moment for prospective evaluation: after the instrument has been finalized but before the resulting human responses exist. At this point, researchers can register model predictions before fielding and score them afterwards. Preregistration therefore acts as a commitment mechanism, fixing predictions before the target outcomes can be observed. The resulting benchmark is free of memorization by construction, and it stays valid as models change, because each new wave supplies questions that no model has seen. The design follows the logic of prediction challenges scored on held-out data (Salganik et al., 2020; Pfänder, 2026). Preregistration also answers calls to constrain the many analytic choices available across data analyses, including studies that use LLMs (Kosch & Feger, 2026; Vaccaro, 2026), and to report them in full (Feuerriegel et al., 2026).

Regarding the difficulty of simulating local knowledge—we find that errors concentrated where an answer depends on a specific street, housing market, or day: how many neighbors a respondent knows, where they would move, whether their windows were open. Simulated respondents also knew too much about places, predicting that almost everyone is familiar with neighborhoods that half of residents know nothing about. Pew Research Center (2026) reports the same excess of knowledge on factual questions. Other surveys should expect simulations to fail on questions of this kind, and future studies should aim to model behaviors, attitudes, and opinions in difficult, localized settings.

Similarly, Ideology was the one axis the zero-shot model could not rank. It also failed in our pilot on the 2024 wave, and the two fine-tuned models disagreed on it. This is notable because ideology is the most salient identity among the axes we tested, and one on which simulated opinion is known to be uneven across groups (Santurkar et al., 2023; Abeliuk et al., 2026). We caution against pre-testing ideology related instruments, especially in localized settings where political dynamics differ from national ones. One possible explanation is that Baltimore is a traditionally liberal city, yielding relatively limited ideological variation in our local sample and making these differences harder to recover. More broadly, ideology may also be less tightly linked to the concrete behaviors and lived circumstances represented in respondents' prior survey answers than axes such as place or employment. These results suggest that the validity of simulation-based pre-testing may depend not only on the question being tested, but also on whether the relevant group distinction is sufficiently represented in the underlying population and respondent context.

Our work is not without limitations. First, we use "pre-testing" in a deliberately narrow sense: rather than evaluating whether simulated respondents can replace cognitive interviewing or identify problems in comprehension, retrieval, response mapping, burden, or item nonresponse, we ask whether they can rank candidate items according to anticipated aggregate response characteristics, such as relative mean responses and subgroup differences. Additionally, successful rank correlation does not by itself establish that simulation-assisted screening improves real questionnaire-selection decisions; future work should therefore test whether these rankings help researchers choose which items to retain when questionnaire space or pre-testing resources are limited, for example through top-k selection, selection regret, or comparisons with simpler baselines and small human pilots.

The study covers one survey, one metropolitan area, one model family, and one pair of waves. We used a single pre-registered prompt and configuration (Dominguez-Olmedo et al., 2024; Ye et al., 2026). The confirmatory tests rest on 22 questions [95% bootstrap interval for means: 0.34 to 0.90], although no single question changes a supported result (Appendix A12). Finally, the candidate questions were written by survey experts for fielding; ranking a battery of weaker, or more similar candidates may be harder.

Lastly, each simulated respondent was built from 258 answers given a year earlier. This richness of persona background likely matters, and is not always feasible. Richer individual information has improved simulations in existing work: long interviews (Park et al., 2024), survey histories (Jia et al., 2026), and in-context survey answers (Holtdirk et al., 2026) all help, while sparse conditioning is where generated data are least realistic (Xie et al., 2026).

In future, the design should be repeated on other surveys and populations, and with other models: fidelity appears to improve with scale, though more slowly for underrepresented opinions (Ziems et al., 2026), and base and post-trained models have different strengths (Grief-Albert et al., 2026). Pre-testing involves more than ranking: future studies can ask whether simulations anticipate item nonresponse, comprehension problems, and the effects of alternative wordings. A further step is to combine simulated rankings with small human pilots, using methods that correct model predictions with human data (Egami et al., 2023; Broska et al., 2025; Krsteski et al., 2026).

## 4. Materials and Methods

### 4.1 Study design

This is a prospective validation study. We used survey data collected from an independent sample in 2025 to initialize simulated respondents, and we compared their predictions for candidate questions against human responses collected in 2026. The unit of analysis is the question. The study was preregistered on July 8, 2026.

**Data collection.** The Baltimore Area Survey (BAS) is an annual survey of Baltimore City and Baltimore County residents, fielded by the 21st Century Cities Initiative. We used the 2025 wave to initialize the simulated respondents (Bader, et al., 2025). The 2025 data are anonymized and publicly available (https://github.com/JHUCities/baltimore-area-survey-data/tree/main/data/bas-

2025). The 2026 wave contains questions new in 2026, that were not asked in 2025. We generated and preregistered predictions for these questions before the 2026 wave was fielded. Human responses to the new questions were collected during the 2026 wave, after preregistration (available at https://osf.io/42cem/overview?view_only=34b11ac21c6941598b3e38c7cea75e3f).

**Questions.** As preregistered, the main confirmatory analysis uses the 22 questions introduced in the 2026 wave. They are answered on a categorical scale and form the pre-tested questions [Transportation: 5; Neighborhood: 7; Connectivity: 6; Housing: 4]. Additional seven questions were registered for exploratory, descriptive analysis: three integer-valued questions (days worked, days worked only from home, household vehicles), three community ratings over 11 named neighborhoods, and one multiple-choice question on types of heating.

**Independent variables.** The independent variable is the value predicted by the simulated respondents: the predicted survey-weighted mean (H1a), the predicted normalized variance (H1b), or the predicted signed regression coefficient for each demographic axis (H1c). All predicted values were generated by zero-shot persona-prompted GPT-4.1 under a fixed configuration. Within the per-question regressions of H1c, the independent variable is the binary demographic indicator.

**Dependent variables.** The dependent variable is the corresponding value observed in the human respondents: the observed normalized survey-weighted mean, the observed normalized variance, or the observed signed regression coefficient. Human responses are recorded on a categorical scale. Within the per-question regressions of H1c, the dependent variable is the response. All 2026 quantities use 2026 survey weights, and all predicted quantities use 2025 survey weights; both weigh to the same target population.

**Model configuration.** Predictions were generated by GPT-4.1 (gpt-4.1-2025-04-14, temperature 0.3), prompting zero-shot with a persona description for each respondent. Each prompt had the same structure: a demographic profile block (all demographic, program, and survey-design variables), followed by a prior-answers block (all substantive 2025 survey responses, excluding analysis and recode variables), followed by the new 2026 questions with their answer options. In total, each simulated respondent was initialized with answers to 258 questions from 2025, which allows in-context learning through prompting. Prompting was done once per simulated respondent.

The exact prompt text, temperature, maximum tokens, and other parameters were selected through pilots on historical data and were frozen before any 2026 responses were collected. GPT-4.1 was selected because it had the highest performance on the pilot data. Appendix A11 lists the complete prompt.

**Sample size.** The pre-tested questions form the sample: 22 questions in the main analysis. Predictions were generated for 1,271 simulated respondents, one for each respondent to the 2025 wave. The number of human respondents was determined by BAS fielding; the 2026 wave collected 1,161 respondents [expected at preregistration: about 1,500].

**Foreknowledge of data.** The predictions were generated and preregistered before any 2026 human responses were received. No 2026 response data were available to the authors or to the model at the time of prediction. The model was initialized only on data from 2025 and earlier. Pilot analyses used to select the model configuration were conducted on historical data (2024 and 2025) and did not use 2026 responses.

### 4.2 Inclusion and exclusion criteria

We included the pre-tested questions specified in the preregistered prediction tables, and human respondents to the 2026 wave who provided a valid response to a given question. For each question, we excluded human responses coded as missing, not presented because of a skip pattern, or opt-out. We excluded simulated responses that could not be parsed into a valid response option for the question; simulated answers were matched to answer options by their label. We required a minimum of 10% valid human responses and 10% valid simulated responses per question. All questions met this threshold [lowest share of valid human responses: 36%, for the three follow-up questions asked only of respondents who had opened their windows].

**Missing data.** The unit of analysis is the question, and each question's mean, variance, and regression coefficient were computed from the available valid responses for that question. We did not impute missing responses. A respondent missing on one question still contributed to other questions. For the H1c regressions, a respondent missing on either the response or the demographic indicator was excluded from that axis's regression for that question. Questions with zero predicted or observed variance were excluded from H1c, because z-scores are not defined [one question: biking frequency, for which every simulated respondent answered "Never"].

### 4.3 Analysis plan

**Means and variances.** For each question we computed the survey-weighted mean of the responses. To make means comparable, answers were scaled to a shared 0–1 scale: normalized mean = (weighted mean − 1) / (number of scale points − 1). For H1b we computed the survey-weighted variance of the responses on the same 0–1 scale.

**Signed group differences.** For each question and each demographic axis, we fitted a survey-weighted simple linear regression of the z-score standardized response on a single binary indicator of the axis, $z = \beta_0 + \beta x + \varepsilon$, where z is the respondent's response standardized by that question's survey-weighted mean and standard deviation, and x is the binary indicator. The coefficient $\beta$ is the signed group difference for that axis and question. Standardizing each question by its own distribution makes $\beta$ comparable across questions with different response scales. Predicted $\beta$ uses the predicted-distribution mean and standard deviation and observed $\beta$ uses the observed-distribution mean and standard deviation, so the two are internally consistent before ranking. The axes were coded as follows: jurisdiction (city = 1, county = 0), race (Black = 1, White = 0; other races excluded), political ideology (conservative or very conservative = 1, liberal or very liberal = 0; moderates excluded), and employment (working full time = 1, unemployed, laid off, or looking for work = 0; all other employment categories excluded). A fifth registered axis, transportation (cannot usually afford transportation = 1, can afford = 0), could not be computed in 2026 (Section 4.6). Group sizes in the 2026 wave were 695 city and 466 county residents, 401 Black and 589 White respondents, 211 conservative and 412 liberal respondents, and 563 full-time employed and 51 unemployed respondents.

**Statistical modelling.** For H1a and H1b, we ranked the questions by the predicted and by the observed quantity and computed the Spearman rank correlation between the two rankings. For H1c, we ranked the questions by the predicted and by the observed coefficient, separately for each axis. All tests are one-sided against the alternative of positive correlation, with a null hypothesis of no rank correlation ($\rho = 0$). H1a and H1b were tested at $\alpha = 0.05$. The H1c tests were tested at $\alpha = 0.05 / 5 = 0.01$, under a Bonferroni correction across the five registered axes.

**Inference criteria and interpretation.** For each hypothesis, we reject the null and conclude that the simulated respondents predict the corresponding quantity if the Spearman rank correlation is positive and significant at the stated threshold. A significant positive correlation indicates that the simulated respondents rank the new questions in the same order as the human respondents on the corresponding quantity. We interpret support for H1a as evidence that the approach can pre-test how questions rank in their answers, H1b as evidence for ranking questions by elicited variance, and H1c as evidence for ranking questions by the signed group difference of a demographic axis. Because the predictions were preregistered before the human responses were collected, a significant result cannot be explained by memorization of the new questions.

**Secondary hypotheses.** The secondary hypotheses were exploratory. For H2, we repeated the H1a, H1b, and H1c analyses using GPT-4.1 fine-tuned on earlier BAS responses in place of the zero-shot model, and compared the rank correlations. For H3, we compared the rank correlations from H1c across axes, together with the share of questions on which the predicted and observed $\beta$ have the same sign and the mean absolute difference between them. For H4, we compared predictive accuracy across the four survey modules, using within-module rank correlations and the mean absolute error of the predicted mean and variance.

### 4.4 Planned additional exploratory analyses

The pre-registration lists planned exploratory analyses. Three further analyses were added after the confirmatory analysis and were not preregistered: bootstrap intervals and leave-one-question-out checks for the rank correlations, the split-half reliability of the human rankings, and baselines built from 2025 responses. Appendix A15 describes the procedures for all twelve analyses, and Appendix A4–A14 the results.

### 4.5 Fine-tuned models and pilot

As a secondary analysis, we repeated all tests with two fine-tuned versions of GPT-4.1, one fine-tuned on 2024 and one on 2025 BAS responses [85 training and 15 validation respondents each], using the same prompt as the zero-shot model. The model and the prompt were selected in a retrospective pilot on the 2024 wave, in which GPT-4.1 outperformed six other models [zero-shot rank correlation for means: $\rho = 0.757$]. Appendix A16 lists the fine-tuning setup and the pilot results.

### 4.6 Deviations from the preregistration

The analysis deviates from the preregistration in four ways. For each, we include a justification and sensitivity to a specific choice. (1) The item that defines the transportation axis ("Can you usually afford the transportation you need?") was not fielded in 2026, so the fifth test of H1c could not be run. (2) Due to a coding error, the code behind the registered prediction tables discarded valid simulated answers on four questions: answers that named the respondent's jurisdiction on the two questions about moving, and some of the five answer labels on the questions about vacant lots and trust in Johns Hopkins University. The main analysis recomputes the predictions from the same stored pre-registered responses, with every valid answer counted. No response was regenerated and the correction uses no 2026 data; all confirmatory tests are therefore also reported with the predictions exactly as registered, and the employment test is the only decision that differs. (3) Due to a coding error, the pre-registered question table describes the item on opening windows as coded Yes = 1, No = 0, whereas the registered predictions were computed with Yes = 0, No = 1; observed values use the coding of the predictions. (4) In 2026, days worked was asked only of employed respondents, while predictions were generated for everyone; we report the observed mean both among those asked and with respondents who were not asked set to zero.

## References

Abeliuk, A., Gaete, V., & Bro, N. (2026). Auditing socio-demographic and cross-societal fairness in LLM-simulated public opinion. *EPJ Data Science, 15*, 75.

Aher, G., Arriaga, R. I., & Kalai, A. T. (2023). Using large language models to simulate multiple humans and replicate human subject studies. *Proceedings of the 40th International Conference on Machine Learning*, PMLR 202.

Anthis, J. R., Liu, R., Richardson, S. M., Kozlowski, A. C., Koch, B., Evans, J., Brynjolfsson, E., & Bernstein, M. (2025). LLM social simulations are a promising research method. arXiv:2504.02234.

Argyle, L. P., Busby, E. C., Fulda, N., Gubler, J. R., Rytting, C., & Wingate, D. (2023). Out of one, many: Using language models to simulate human samples. *Political Analysis, 31*(3), 337–351.

Ashokkumar, A., Hewitt, L., Ghezae, I., & Willer, R. (2026). Large Language Models Can Predict the Results of Social Science Experiments. *Nature.*

Bader, Michael DM, Mac MacComas, Alexi Williams, Marin Beal, Leslie Carson, Kabir Broker, Rachel Oyewo, Sarah DeGennaro, Mia Schildbach. 2025. "Baltimore Area Survey 2025." Version 1.0. December 19, 2025. https://github.com/JHUCities/baltimore-area-survey-data/.

Bail, C. A. (2024). Can generative AI improve social science? *Proceedings of the National Academy of Sciences, 121*(21), e2314021121.

Bisbee, J., Clinton, J. D., Dorff, C., Kenkel, B., & Larson, J. M. (2024). Synthetic replacements for human survey data? The perils of large language models. *Political Analysis, 32*(4), 401–416.

Broska, D., Howes, M., & van Loon, A. (2025). The mixed subjects design: Treating large language models as potentially informative observations. *Sociological Methods & Research*.

McColl, E. (2006). Cognitive interviewing. A tool for improving questionnaire design.

Cheng, M., Piccardi, T., & Yang, D. (2023). CoMPosT: Characterizing and evaluating caricature in LLM simulations. *Proceedings of the 2023 Conference on Empirical Methods in Natural Language Processing*.

Cui, Z., Li, N., & Zhou, H. (2025). A large-scale replication of scenario-based experiments in psychology and management using large language models. *Nature Computational Science, 5*(8), 627–634.

Egami, N., Hinck, M., Stewart, B. M., & Wei, H. (2023). Using imperfect surrogates for downstream inference: Design-based supervised learning for social science applications of large language models. *Advances in Neural Information Processing Systems, 36*.

Feuerriegel, S., Barrie, C., Crockett, M. J., Globig, L. K., McLoughlin, K. L., Mirea, D.-M., … Horta Ribeiro, M. (2026). A reporting checklist for large language models in behavioural science. *Nature Human Behaviour, 10*(7), 1182–1186.

Gelman, Andrew, and Eric Loken. 2013. "The Garden of Forking Paths: Why Multiple Comparisons Can Be a Problem, Even When There Is No 'Fishing Expedition' or 'p-Hacking' and the Research Hypothesis Was Posited Ahead of Time." Open Science Foundation.

Grief-Albert, S., Bo, J., Jiao, D., & Anderson, A. (2026). Emulate or estimate? The divergent strengths of base and post-trained language models for opinion simulation. arXiv:2608.03044.

Hazra, S., Majumder, B. P., Siangliulue, P., Clark, P., Zhang, A. X., & Chang, J. (2026). UnexpectedParrots: Iterating Behavioral Experiment Design with In-Silico Simulations. arXiv preprint arXiv:2610.25343.

Holtdirk, T., Ahnert, G., Sakshaug, J. W., & Haensch, A.-C. (2026). In-context learning for the imputation of public opinion data with large language models. arXiv:2606.09351.

Horton, J. J. (2023). Large language models as simulated economic agents: What can we learn from Homo Silicus? NBER Working Paper No. 31122.

Hu, T., Baumann, J., Lupo, L., Collier, N., Hovy, D., & Röttger, P. (2026). SimBench: Benchmarking the ability of large language models to simulate human behaviors. *International Conference on Learning Representations*.

Hullman, J., Broska, D., Sun, H., & Shaw, A. (2026). This human study did not involve human subjects: Validating LLM simulations as behavioral evidence. arXiv:2602.15785.

Jia, M., Chen, Y., Sharma, D., & Diaz-Rodriguez, J. (2026). When can digital personas reliably approximate human survey findings? arXiv:2605.10659.

Kalai, A. T., Nachum, O., Vempala, S. S., & Zhang, E. (2026). Evaluating large language models for accuracy incentivizes hallucinations. Nature, 1-3.

Kosch, T., & Feger, S. (2026). Prompt-hacking: The new p-hacking? *Communications of the ACM, 69*(3), 35–37.

Krsteski, S., Russo, G., Chang, S., West, R., & Gligorić, K. (2026). Valid survey simulations with limited human data: The roles of prompting, fine-tuning, and rectification. Proceedings of the 64th Annual Meeting of the Association for Computational Linguistics.

Ludwig, J., Mullainathan, S., & Rambachan, A. (2025). Large language models: An applied econometric framework. NBER Working Paper No. 33344.

Ma, X., Zhang, M., Ang, S., & Chen, M. (2026). Tastes without distinction: Silicon samples and the synthetic construction of tastes. arXiv:2606.30085.

Nosek, B. A., Ebersole, C. R., DeHaven, A. C., & Mellor, D. T. (2018). The preregistration revolution. Proceedings of the National Academy of Sciences, 115(11), 2600-2606.

Olivos, F., & Liu, M. (2024). ChatGPTest: Opportunities and cautionary tales of utilizing AI for questionnaire pretesting. *Field Methods*. https://doi.org/10.1177/1525822X241280574

Park, J. S., Zou, C. Q., Shaw, A., Hill, B. M., Cai, C., Morris, M. R., Willer, R., Liang, P., & Bernstein, M. S. (2024). Generative agent simulations of 1,000 people. arXiv:2411.10109.

Perez, C. J., Vasquez Jr., M. P., Giabbanelli, P., & Wu, P. Y. (2026). Simulating social attitudes with LLMs: Accuracy, demographic effects, and refusal behavior in the sensitive domain of suicide prevention. *Proceedings of the Seventh Workshop on Natural Language Processing and Computational Social Science*, 176–189.

Pew Research Center. (2026, September 30). *Can AI stand in for human survey-takers? Not really*. https://www.pewresearch.org/data-labs/2026/09/30/can-ai-stand-in-for-human-survey-takers-not-really/

Pfänder, J. (2026). *The Silicon Sample Benchmark: Predicting a megastudy before the results are revealed*. https://janpfander.github.io/llm_predictions_megastudy/

Romberg, J., Gummer, T., Lapesa, G., Kunz, T., & Wagner, C. (2026). Hybrid Panels: Toward Human-AI Collaboration in Survey Research. arXiv preprint arXiv:2608.22582.

Rothschild, D. M., Buskirk, T. D., Eckman, S., Hillygus, D. S., Kreuter, F., & Lazer, D. (2025). Successfully navigating the disruption AI will bring to survey research. The Survey Statistician, 92(1), 30-44.

Rothschild, D., Marlar, J., Amaya, A., Barari, S., Buskirk, T., Cobb, C., Gennai, J., Hillygus, D., Korlakai Vinayak, R., Krupenkin, M., Lee, S., Steiger, D., & Webb, B. (2026). *Responsible AI integration in survey research*. American Association for Public Opinion Research.

Presser, S., Couper, M. P., Lessler, J. T., Martin, E., Martin, J., Rothgeb, J. M., & Singer, E. (2004). Methods for testing and evaluating survey questions. Methods for testing and evaluating survey questionnaires, 1-22.

Salganik, M. J., Lundberg, I., Kindel, A. T., Ahearn, C. E., Al-Ghoneim, K., Almaatouq, A., … McLanahan, S. (2020). Measuring the predictability of life outcomes with a scientific mass collaboration. *Proceedings of the National Academy of Sciences, 117*(15), 8398–8403.

Santurkar, S., Durmus, E., Ladhak, F., Lee, C., Liang, P., & Hashimoto, T. (2023). Whose opinions do language models reflect? *Proceedings of the 40th International Conference on Machine Learning*, 29971–30004.

Schaeffer, Nora Cate, and Stanley Presser. 2003. "The Science of Asking Questions." Annual Review of Sociology 29(1):65–88. doi:10.1146/annurev.soc.29.110702.110112.

Sturgis, P., Roberts, C., & Robinson, T. (2026). LLMs for survey pretesting: How well can large language models identify flaws in survey questions? Survey Futures Working Paper No. 15. University of Essex.

Suh, J., Jahanparast, E., Moon, S., Kang, M., & Chang, S. (2025). Language model fine-tuning on scaled survey data for predicting distributions of public opinions. *Proceedings of the 63rd Annual Meeting of the Association for Computational Linguistics*, 21147–21170.

Vaccaro, M. (2026). Preregistration for experiments with AI agents. arXiv:2606.11217.

Wang, A., Morgenstern, J., & Dickerson, J. P. (2025). Large language models that replace human participants can harmfully misportray and flatten identity groups. *Nature Machine Intelligence, 7*(3), 400–411.

Xie, Y., Liang, L., Li, S., Lu, Y., Xiao, Z., Shi, M., Huang, J., Wang, M., & Xie, Y. (2026). Evaluating the statistical realism of LLM-generated social science data. *Proceedings of the National Academy of Sciences, 123*(19), e2538145123.

Xu, F., Shao, C., Chen, L., Zeng, Q., Chen, Z., Sukiennik, N., ... & Evans, J. (2026). Large language models for computational social science. Nature Human Behaviour, 1-17.

Ye, J., Cao, L., Chen, D., & Ferrara, E. (2026). Stop drawing scientific claims from LLM social simulations without robustness audits. arXiv:2605.18890.

Yu, S.-Z., Suh, J., Chang, S., & Chan, D. M. (2026). Anamnesis: An open-source platform for large-scale backstory-conditioned survey simulation. arXiv:2607.10628.

Ziems, C., Held, W., Karaca, S. D., Grusky, D., Hashimoto, T., & Yang, D. (2026). Will scaling improve social simulation with LLMs? arXiv:2607.02464.

# Can LLM-Simulated Respondents Pre-Test Survey Questions? Evidence From Preregistered Prospective Validation

## Appendix

### A1. Predictions and observations for every question

Tables A1 and A2 list the zero-shot prediction and the 2026 observation for each of the 22 pre-tested questions and for every quantity tested. The six transportation-insecurity items that were also asked in 2025 are reported separately in Appendix A5.

| Question | Module | Mean: pred. | Mean: obs. | Var.: pred. | Var.: obs. |
|---|---|---|---|---|---|
| Biking frequency past 30 days | Transportation | 1.00 | 0.95 | 0.00 | 0.03 |
| Own vehicle frequency past 30 days | Transportation | 0.26 | 0.29 | 0.17 | 0.14 |
| Public transit frequency past 30 days | Transportation | 0.90 | 0.87 | 0.05 | 0.07 |
| Transport trouble for medical appointments | Transportation | 0.14 | 0.09 | 0.09 | 0.06 |
| Walking frequency past 30 days | Transportation | 0.30 | 0.67 | 0.02 | 0.13 |
| How long windows were open | Housing | 0.58 | 0.61 | 0.03 | 0.13 |
| How wide windows were opened | Housing | 0.43 | 0.49 | 0.03 | 0.12 |
| Largest number of windows open at once | Housing | 0.30 | 0.36 | 0.01 | 0.10 |
| Windows open in past 24 hours | Housing | 0.36 | 0.62 | 0.23 | 0.23 |
| Business influence on community | Connectivity | 0.51 | 0.51 | 0.03 | 0.06 |
| Business meets family needs | Connectivity | 0.22 | 0.18 | 0.07 | 0.06 |
| Business reflects preferences | Connectivity | 0.49 | 0.55 | 0.04 | 0.06 |
| Business satisfaction | Connectivity | 0.55 | 0.55 | 0.03 | 0.07 |
| Fear of deportation | Connectivity | 0.02 | 0.16 | 0.01 | 0.08 |
| Trust in Johns Hopkins University | Connectivity | 0.56 | 0.62 | 0.02 | 0.07 |
| Destination if moving | Neighborhood | 0.28 | 0.57 | 0.06 | 0.10 |
| Ease of finding equivalent home | Neighborhood | 0.59 | 0.40 | 0.03 | 0.09 |
| Likelihood of moving in 3 years | Neighborhood | 0.22 | 0.34 | 0.09 | 0.12 |
| Neighbors known by name | Neighborhood | 0.43 | 0.62 | 0.04 | 0.06 |
| Preferred residence in 3 years | Neighborhood | 0.20 | 0.50 | 0.12 | 0.15 |
| Vacant lots as a problem | Neighborhood | 0.16 | 0.13 | 0.07 | 0.06 |
| Worry about affording to stay | Neighborhood | 0.37 | 0.29 | 0.13 | 0.10 |

**Table A1.** Predicted (zero-shot) and observed mean and variance of answers, on a 0–1 scale.

| Question | City: pred. | City: obs. | Race: pred. | Race: obs. | Ideol.: pred. | Ideol.: obs. | Empl.: pred. | Empl.: obs. |
|---|---|---|---|---|---|---|---|---|
| Biking frequency past 30 days | — | −0.22 | — | −0.16 | — | 0.02 | — | 0.14 |
| Own vehicle frequency past 30 days | 0.70 | 0.42 | 0.76 | 0.39 | −0.08 | −0.16 | −1.12 | −1.24 |
| Public transit frequency past 30 days | −0.58 | −0.51 | −0.62 | −0.41 | 0.25 | 0.14 | 0.59 | 0.73 |
| Transport trouble for medical appointments | 0.46 | 0.35 | 0.61 | 0.32 | −0.00 | 0.02 | −0.95 | −0.56 |
| Walking frequency past 30 days | −0.38 | −0.49 | −0.50 | −0.14 | 0.32 | 0.25 | 0.32 | 0.47 |

| Question | City: pred. | City: obs. | Race: pred. | Race: obs. | Ideol.: pred. | Ideol.: obs. | Empl.: pred. | Empl.: obs. |
|---|---|---|---|---|---|---|---|---|
| How long windows were open | 0.07 | 0.41 | −0.41 | 0.11 | −0.35 | 0.06 | 0.29 | −0.11 |
| How wide windows were opened | 0.01 | −0.23 | −0.03 | −0.63 | −0.20 | −0.07 | −0.21 | 0.13 |
| Largest number of windows open at once | −0.35 | −0.05 | −0.71 | −0.23 | −0.18 | 0.60 | 0.69 | −0.30 |
| Windows open in past 24 hours | 0.02 | −0.14 | −0.01 | −0.57 | −0.01 | −0.02 | −0.13 | 0.22 |
| Business influence on community | −0.55 | −0.34 | −0.26 | −0.43 | −0.06 | −0.00 | 0.87 | 0.14 |
| Business meets family needs | 0.58 | 0.62 | 0.66 | 0.43 | 0.00 | −0.13 | −1.28 | −0.59 |
| Business reflects preferences | −0.68 | −0.55 | −0.41 | −0.42 | −0.01 | 0.19 | 0.94 | 0.29 |
| Business satisfaction | −0.65 | −0.48 | −0.31 | −0.35 | −0.04 | 0.13 | 0.95 | 0.06 |
| Fear of deportation | −0.07 | 0.04 | −0.09 | 0.17 | 0.02 | −0.25 | 0.14 | 0.06 |
| Trust in Johns Hopkins University | −0.29 | −0.32 | −0.42 | −0.40 | −0.45 | −0.12 | 0.80 | 0.02 |
| Destination if moving | 0.74 | 0.11 | −0.00 | −0.19 | −0.11 | 0.40 | −0.35 | 0.28 |
| Ease of finding equivalent home | 0.31 | 0.03 | 0.29 | −0.00 | −0.06 | 0.04 | −0.61 | −0.30 |
| Likelihood of moving in 3 years | 0.52 | 0.24 | 0.58 | 0.25 | −0.13 | −0.01 | −1.31 | −0.27 |
| Neighbors known by name | 0.61 | 0.06 | 0.49 | 0.41 | −0.06 | −0.05 | −1.07 | −0.03 |
| Preferred residence in 3 years | 0.54 | 0.14 | 0.28 | 0.16 | −0.06 | 0.37 | −1.01 | 0.11 |
| Vacant lots as a problem | 0.94 | 0.53 | 0.55 | 0.38 | −0.07 | −0.09 | −0.68 | −0.64 |
| Worry about affording to stay | 0.37 | 0.02 | 0.62 | 0.08 | −0.23 | −0.04 | −0.75 | −0.14 |

**Table A2.** Predicted (zero-shot) and observed signed group differences (β, in standard deviations) for jurisdiction, race, ideology, and employment.

## A2. Sensitivity to the registered parsing

Table A3 reports every confirmatory test under both ways of reading the simulated responses: the corrected parsing used in the main analysis, and the parsing behind the registered prediction tables, which discarded valid answers on four questions (Section 4.6). For the zero-shot model, the registered column uses the values exactly as printed in the preregistration.

One decision differs: employment for the zero-shot model [$\rho = 0.57$, $p = 0.003$ corrected; $\rho = 0.43$, $p = 0.026$ registered]. Every other decision, for all three models, is the same (Table A3).

| Test | Zero-shot: corrected | Zero-shot: registered | FT-25: corrected | FT-25: registered | FT-24: corrected | FT-24: registered |
|---|---|---|---|---|---|---|
| H1a: mean answer | 0.71* | 0.62* | 0.69* | 0.70* | 0.64* | 0.62* |
| H1b: variance | 0.31 | 0.29 | 0.44* | 0.50* | 0.43* | 0.44* |
| H1c-1: city vs. county | 0.88* | 0.88* | 0.89* | 0.87* | 0.75* | 0.78* |
| H1c-2: Black vs. White | 0.68* | 0.68* | 0.71* | 0.71* | 0.66* | 0.65* |
| H1c-3: conservative vs. liberal | 0.08 | 0.36 | 0.11 | 0.13 | 0.54* | 0.56* |
| H1c-4: full-time vs. unemployed | 0.57* | 0.43 | 0.57* | 0.54* | 0.55* | 0.50* |

**Table A3.** Spearman ρ on the 22 main questions under the corrected and the registered parsing. * H0 rejected at the registered threshold ($\alpha = 0.05$ for H1a and H1b; $\alpha = 0.01$ for H1c). FT-25 and FT-24: GPT-4.1 fine-tuned on 2025 and on 2024 data.

### A3. Fine-tuned models

Figure A1 compares the three models on means and variances. The fine-tuned models rank means about as well as the zero-shot model [ρ = 0.69 and 0.64 vs. 0.71] and rank variances better [ρ = 0.44 and 0.43 vs. 0.31]. Their predicted variances are larger and closer to the observed ones, which is where the gain comes from [mean variance: 0.09 and 0.09 fine-tuned vs. 0.06 zero-shot; 0.10 observed] (Figure A1).

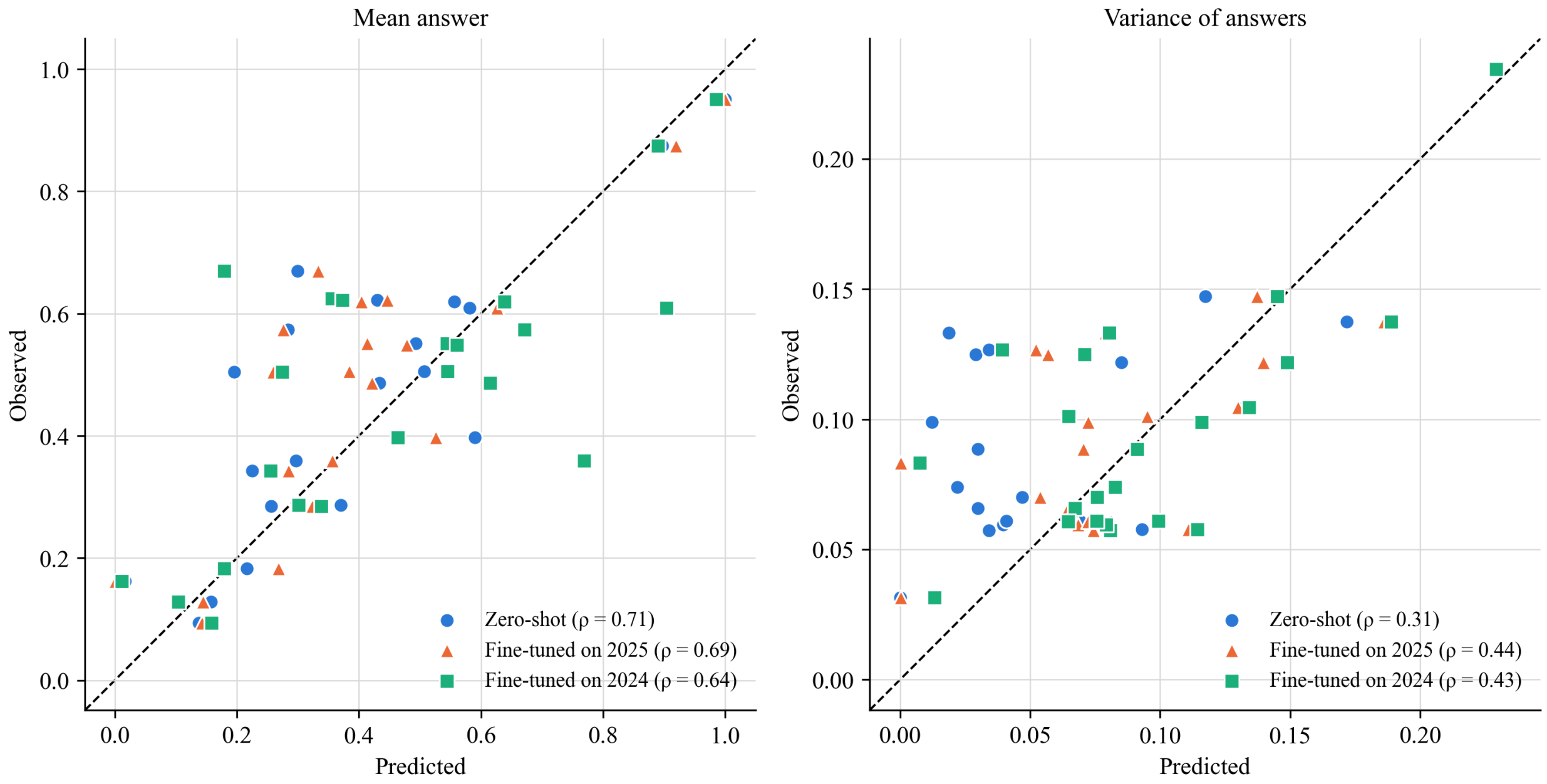

**Figure A1.** Predicted and observed means (left) and variances (right) for the 22 main questions, for the zero-shot model and the two fine-tuned models. The dashed diagonal marks equality.

### A4. Full answer distributions

A two-sample Kolmogorov–Smirnov test compares the simulated and the observed answers to each question. Equality is rejected for almost all of the 22 questions: it is not rejected for 2 questions with the zero-shot model [transport trouble for medical appointments and public transit frequency], for 1 with the model fine-tuned on 2025 data [vacant lots], and for none with the model fine-tuned on 2024 data [α = 0.05]. The six transportation-insecurity items repeated from 2025 pass for every model (Appendix A5).

Figure A2 shows the five questions furthest from the human distribution. Simulated answers concentrate on one or two options while people spread across the scale [walking “a few times a week”: 85% simulated vs. 16% observed; “stay in your current home”: 75% vs. 30%]. Fine-tuning narrows the gap for ease of finding a home [distance 0.46 zero-shot vs. 0.21–0.25 fine-tuned] but not for walking [0.63 vs. 0.54 and 0.57] or neighbors known [0.34 vs. 0.42 and 0.53].

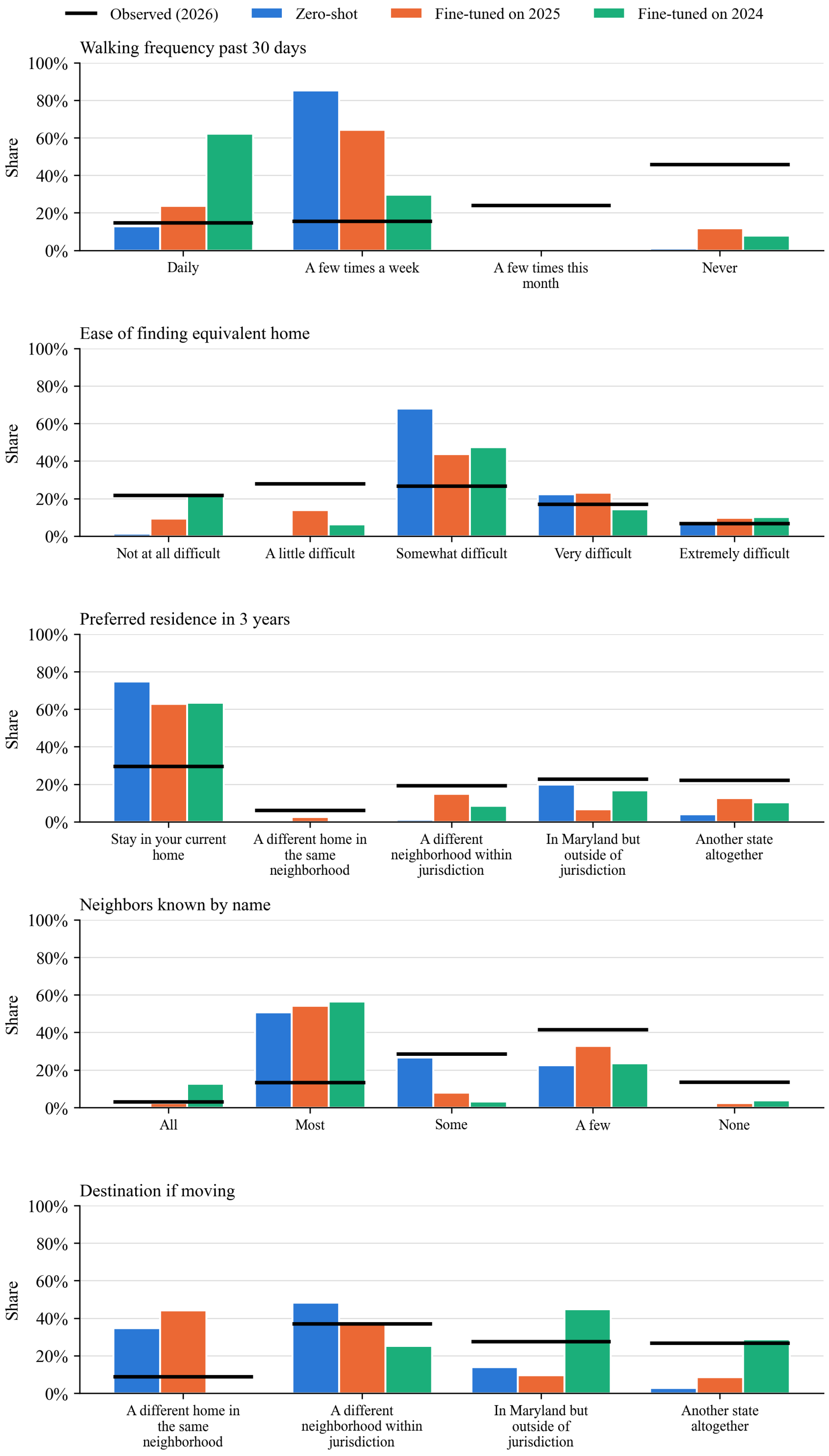

**Figure A2.** Simulated and observed answer distributions for the five questions with the largest zero-shot Kolmogorov–Smirnov distance. Bars: weighted share of simulated answers for each model. Black line: observed 2026 share.

### A5. Six transportation-insecurity items repeated from 2025

The preregistration listed 28 pre-tested questions and designated six of them, on transportation insecurity, as a sensitivity analysis rather than part of the main set, because they overlap with items each simulated respondent had answered in 2025. The preregistration described them as related but not identical to those items. Their titles differ, but the question wording and the three answer options [often, sometimes, never] are identical to the 2025 items. The same items were also asked in 2023 and 2024.

Because each respondent's 2025 answers to these items were in the prompt, the simulated respondents did not predict them: for all three models, every simulated respondent with a valid 2025 answer gave exactly that answer again. The predicted means therefore equal the 2025 means, and the 2026 means hardly changed [differences of 0.00 to 0.03; Table A4]. All six items pass the Kolmogorov–Smirnov test for every model (Appendix A4).

| Question | 2025 variable | Similarity | Predicted mean | 2025 mean | 2026 mean |
|---|---|---|---|---|---|
| Could not leave house due to transport | tsp_noleave | 0.85 | 0.85 | 0.85 | 0.85 |
| Worried about inconveniencing others re transport | tsp_inconv | 0.78 | 0.82 | 0.82 | 0.83 |
| Skipped going somewhere due to transport | tsp_skipgo | 0.73 | 0.84 | 0.84 | 0.84 |
| Rescheduled appointment due to transport | tsp_appt | 0.73 | 0.88 | 0.88 | 0.85 |
| Transport problems affected relationships | tsp_affrel | 0.72 | 0.88 | 0.88 | 0.86 |
| Felt down due to transport problems | tsp_feeldown | 0.64 | 0.82 | 0.82 | 0.84 |

**Table A4.** The six transportation-insecurity items: predicted mean (zero-shot), observed 2025 mean, and observed 2026 mean, on a 0–1 scale. The fine-tuned models give the same predicted means. Similarity: TF-IDF cosine similarity between the 2026 and the 2025 question text (Appendix A14).

As preregistered, we also recomputed the confirmatory tests on all 28 questions. Adding the six items raises most correlations for the zero-shot model [means: $\rho = 0.86$; variance: 0.38, $p = 0.024$; jurisdiction: 0.90; race: 0.64; ideology: 0.17; employment: 0.64], and the fine-tuned models show the same pattern [fine-tuned on 2025: 0.85, 0.49, 0.91, 0.70, 0.09, 0.69; fine-tuned on 2024: 0.80, 0.43, 0.81, 0.67, 0.53, 0.67, in the same order]. The only decision that changes is variance, which passes on 28 questions (Table A5). Because the six items reproduce answers rather than predict them, the 28-question results overstate predictive accuracy, and the 22 questions are the appropriate test. One question in the main set, whether windows were open in the past 24 hours, was also asked in 2025 and was reproduced in the same way; removing it changes no decision (Appendix A14).

| Test | Zero-shot: 22 | Zero-shot: 28 | FT-25: 22 | FT-25: 28 | FT-24: 22 | FT-24: 28 |
|---|---|---|---|---|---|---|
| H1a: mean answer | 0.71 | 0.86 | 0.69 | 0.85 | 0.64 | 0.80 |
| H1b: variance | 0.31 | 0.38 | 0.44 | 0.49 | 0.43 | 0.43 |
| H1c-1: city vs. county | 0.88 | 0.90 | 0.89 | 0.91 | 0.75 | 0.81 |
| H1c-2: Black vs. White | 0.68 | 0.64 | 0.71 | 0.70 | 0.66 | 0.67 |
| H1c-3: conservative vs. liberal | 0.08 | 0.17 | 0.11 | 0.09 | 0.54 | 0.53 |
| H1c-4: full-time vs. unemployed | 0.57 | 0.64 | 0.57 | 0.69 | 0.55 | 0.67 |

**Table A5.** Spearman ρ on the 22 main questions and on all 28 pre-tested questions, for the three models (21 and 27 questions for H1c with the zero-shot model). FT-25 and FT-24: GPT-4.1 fine-tuned on 2025 and on 2024 data.

## A6. Robustness of the variance ranking

On a bounded scale the variance depends on the mean and on the number of answer options. Restricted to the 12 questions that share a 1–5 scale, the variance ranking disappears [$\rho = 0.03$, $p = 0.47$].

Expressed relative to its maximum given the mean, variance / ($m(1 - m)$), the variance is ranked well [$\rho = 0.66$, $p < 0.001$, $n = 21$]. The two checks point in different directions (Figure A3): the model captures how dispersed answers are for a given mean, and misses differences among questions with the same format. A baseline built from 2025 responses ranks variance better than the model [$\rho = 0.43$ vs. 0.31; Table A11].

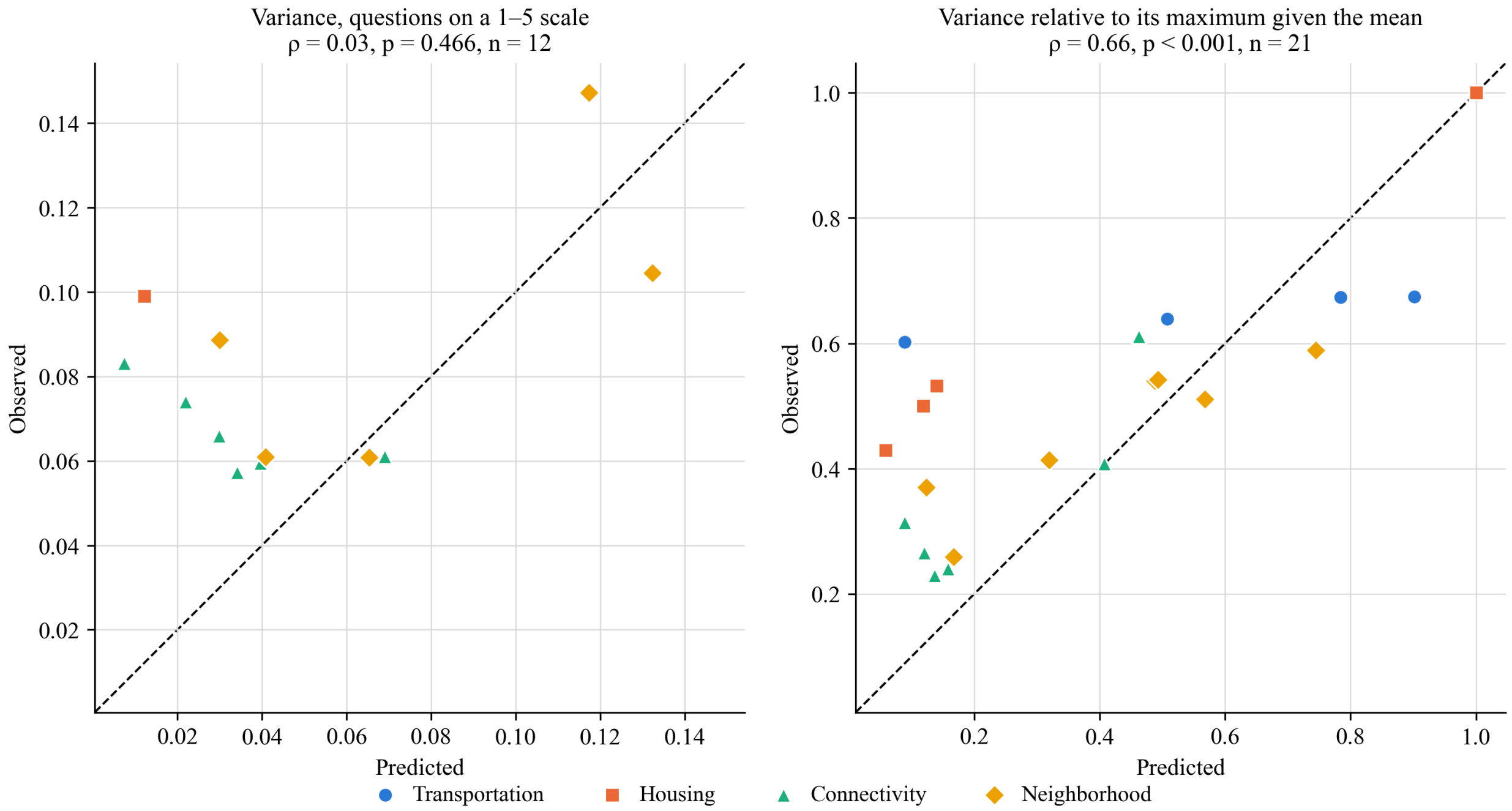

**Figure A3.** Predicted and observed variance for the 12 main questions on a 1–5 scale (left), and variance relative to its maximum given the mean for the main questions (right; biking frequency excluded). Zero-shot model.

## A7. Community ratings, heating types, and counts

Three questions ask which of 11 neighborhoods respondents do not know, would never consider, or would seriously consider. The desirability ordering stated in the preregistration matches the neighborhoods residents would seriously consider [$\rho = 0.88$] and would never consider [$\rho = -0.85$]. Predicted shares order the neighborhoods well for "seriously consider" [0.90 zero-shot; 0.86 and 0.79 fine-tuned], moderately for "never consider" [0.51; 0.33 and 0.74], and poorly for "do not know" [−0.36; 0.20 and 0.34]. The zero-shot model predicts that 4% of residents know nothing about Homeland, against 56% observed (Figure A4).

The order of the seven heating types is predicted well [$\rho = 0.82$ zero-shot; 0.86 and 0.96 fine-tuned], with central heating and space heaters over-predicted [zero-shot: 90% vs. 77% observed; 22% vs. 10%]. These analyses are descriptive and untested [$n = 11$ and $n = 7$].

Counts are poorly calibrated, as the preregistration expected (Table A6). With the population matched to the prediction, predicted days worked fall inside the observed 90% interval [5.6 vs. 5.5; CI 5.1–5.9]. Days worked only from home [1.6 vs. 2.7; CI 2.4–3.1] and household vehicles [1.3 vs. 2.3; CI 2.0–2.6] are under-predicted by all three models.

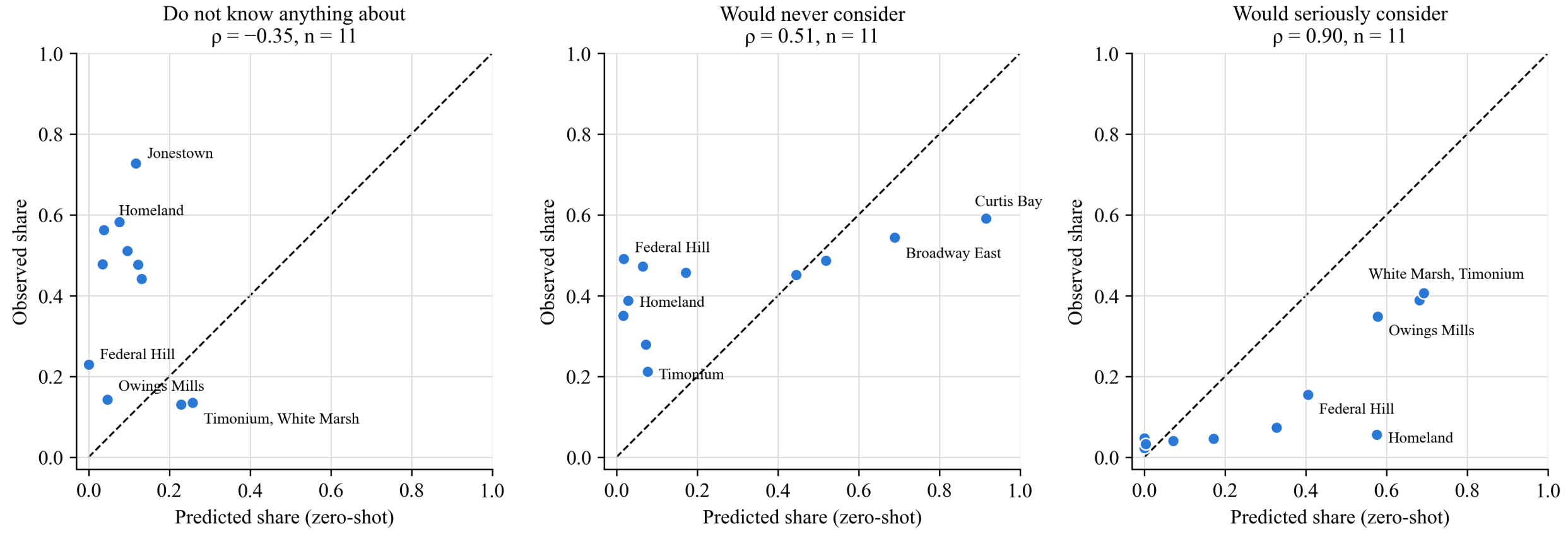

**Figure A4.** Community ratings: predicted (zero-shot) and observed share of respondents selecting each of 11 neighborhoods, for the three items. The dashed diagonal marks equality; a few neighborhoods are named.

| Question | Observed mean (90% CI) | Zero-shot | FT-25 | FT-24 |
|---|---|---|---|---|
| Days worked, past 14 days (those asked) | 9.6 (9.2 to 9.9) | 5.6 | 5.6 | 6.0 |
| Days worked, past 14 days (all; 0 if not asked) | 5.5 (5.1 to 5.9) | 5.6 | 5.6 | 6.0 |
| Days worked only from home | 2.7 (2.4 to 3.1) | 1.6 | 0.3 | 1.2 |
| Household vehicles | 2.3 (2.0 to 2.6) | 1.3 | 1.2 | 1.3 |

**Table A6.** Integer-valued questions: observed weighted mean with 90% bootstrap interval, and predicted weighted mean for each model. Days worked was asked only of employed respondents in 2026; the second row sets those not asked to zero.

## A8. Year-to-year stability of human answers

Forty-four questions were asked in both waves with the same answer codes. Among human respondents, the 2025 ranking almost reproduces the 2026 ranking [means: $\rho = 0.98$; variances: $\rho = 0.89$].

The largest shifts are lower financial worry in 2026 [normalized means: food 0.35 to 0.23; medical costs 0.40 to 0.30; housing 0.34 to 0.24] and fewer respondents who cannot rely on their vehicle [21% in 2025 vs. 8% in 2026]. Drift between waves is small relative to the prediction error on new questions, so it does not explain the gap [means: $\rho = 0.98$ between waves vs. 0.71 between the model and respondents]. The repeated questions are a different set from the new ones, so this is a benchmark and not a decomposition (Figure A5).

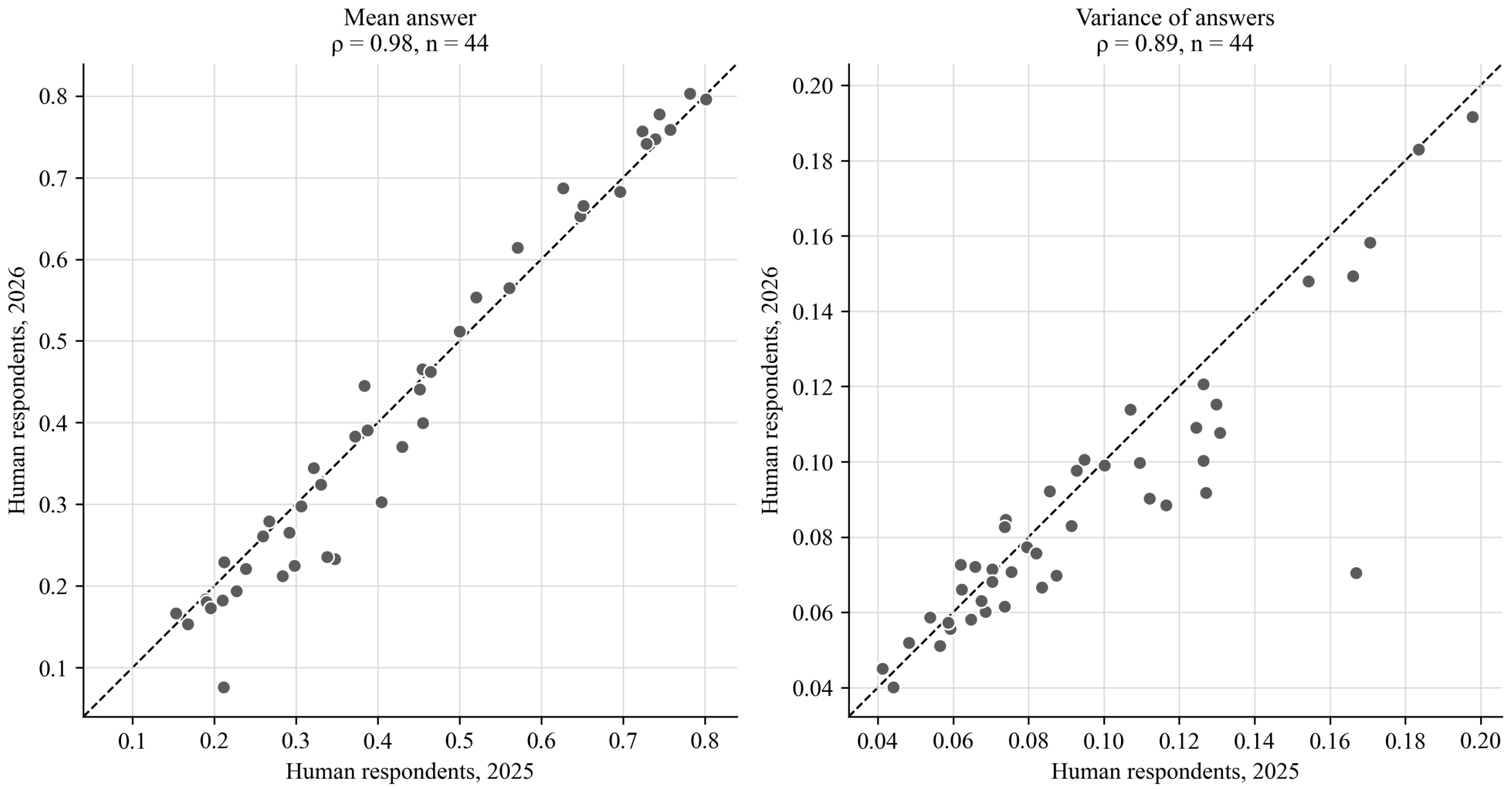

**Figure A5.** Mean (left) and variance (right) of human answers in 2025 and in 2026 for the 44 questions asked in both waves. Each point is a question; the dashed diagonal marks no change.

## A9. Modules and question types

Accuracy differs by module [mean absolute error of the predicted mean: Connectivity 0.05, Housing 0.10, Transportation 0.10, Neighborhood 0.17; within-module rank correlation of means: 0.83, 0.40, 1.00, 0.43]. In Transportation, the error comes almost entirely from walking frequency [0.30 predicted vs. 0.67 observed]; the other four questions are within 0.05. Connectivity ranks means well but not variances [$\rho = -0.77$], and the Housing questions often have group differences in the wrong direction [employment: 0 of 4 signs correct] (Table A7).

Under our own classification, behaviors and experiences are ranked better than attitudes and perceptions on means [$\rho = 0.85$, $n = 9$ vs. $\rho = 0.63$, $n = 13$] and on variance [0.53 vs. 0.40]. Race differences are ranked better for attitudes [0.80 vs. 0.41]; jurisdiction and employment are similar for both [0.87 vs. 0.91; 0.60 vs. 0.57].

Each module and type contains few questions, so these comparisons are descriptive (Table A7).

| Group | Questions | ρ: means | ρ: variances | MAE: mean |
|---|---|---|---|---|
| Module: Transportation | 5 | 1.00 | 0.60 | 0.10 |
| Module: Connectivity | 6 | 0.83 | −0.77 | 0.05 |
| Module: Housing | 4 | 0.40 | 1.00 | 0.10 |
| Module: Neighborhood | 7 | 0.43 | 0.64 | 0.17 |

| Group | Questions | ρ: means | ρ: variances | MAE: mean |
|---|---|---|---|---|
| Type: behaviors and experiences | 9 | 0.85 | 0.53 | 0.10 |
| Type: attitudes and perceptions | 13 | 0.63 | 0.40 | 0.12 |

**Table A7.** Zero-shot accuracy by survey module and by question type, 22 main questions. ρ: Spearman correlation between predicted and observed values within the group. MAE: mean absolute error of the predicted mean on the 0–1 scale.

## A10. Signed and absolute group differences

The preregistered tests rank questions by the signed difference between groups. Ranking by the absolute difference asks a harder question: which questions separate the groups most, regardless of direction.

The model answers it for jurisdiction only [$|\beta|$: $\rho = 0.48$, $p = 0.014$; race: −0.05; ideology: 0.12; employment: 0.07; $n = 21$]. The signed correlations for race and employment therefore come from getting the direction of the difference right across questions, and not from anticipating where the gap is largest (Table A8).

| Axis | Signed β: ρ | Signed β: p | \|β\|: ρ | \|β\|: p |
|---|---|---|---|---|
| City vs. county | 0.88 | < 0.001 | 0.48 | 0.014 |
| Black vs. White | 0.68 | < 0.001 | −0.05 | 0.580 |
| Conservative vs. liberal | 0.08 | 0.358 | 0.12 | 0.301 |
| Full-time vs. unemployed | 0.57 | 0.003 | 0.07 | 0.377 |

**Table A8.** Spearman ρ between predicted and observed group differences, signed and in absolute value, with one-sided p-values. Zero-shot model, 21 questions.

## A11. Prompt

Every prompt opened with “You are a resident of Baltimore answering a local survey.” It then gave two blocks, “Here is your profile:” and “Here are your previous survey answers:”. Each entry showed a short title, the question text, the answer options, and that respondent’s 2025 answer.

The new questions followed under this instruction: “For each question below, give only a short direct answer (the answer text only, no explanation). Number your answers to match the questions. For questions asking you to select multiple items, separate them with ‘ | ’. For number questions, write only the number.” The new questions were listed in a fixed order, each with its title, text, and answer options.

Answers were matched to questions by number and to answer options by label. The fine-tuned models received the same prompt (Section 4.5). Box A1 shows an abridged prompt.

```
You are a resident of Baltimore answering a local survey.

Here is your profile:
  [Home Owner Or Renter] Q: "Do you rent your home, own it, or do you have some other arrangement?
     Options: Rent | Own | Some other arrangement"
  A: Own
  [...]

Here are your previous survey answers:
  [Drug Use In Neighborhood] Q: "How much of a problem is the use of illegal illicit drugs in your
neighborhood?
     Options: Not a problem at all | A small problem | Somewhat of a problem | A big problem | A
major problem"
  A: A small problem
  [...]

=== PLEASE ANSWER THESE QUESTIONS ===
For each question below, give only a short direct answer (the answer text only, no explanation).
Number your answers to match the questions.
For questions asking you to select multiple items, separate them with " | ".
For number questions, write only the number.

  1. [Neighbors known by name] Thinking about the people who live on your street. How many of them
do you think you know by name?
     Options: All | Most | Some | A few | None
  [...]
  35. [Transport trouble for medical appointments] In the past 12 months, that is since September
2025 how often did you have trouble getting transportation to medical appointments?
     Options: Never | Once or twice | Three or four times | More than four times
```

**Box A1.** Abridged prompt for one simulated respondent. [...] marks omitted entries: the profile and previous-answer blocks together hold that respondent's 2025 answers, and new questions follow the instruction.

### A12. Uncertainty on the rank correlations

This analysis was not preregistered. For each test we resampled questions with replacement [10,000 resamples] to obtain a 95% interval for ρ, and recomputed the test with each question removed in turn (Table A9; Figure A6).

The four supported zero-shot results do not depend on any single question, and their intervals exclude zero (Table A9). The intervals are wide: only jurisdiction is pinned down as a strong correlation [0.67 to 0.95, against 0.34 to 0.90 for means]. The variance result is the least stable, moving across the threshold when any of three questions is removed [leave-one-out ρ from 0.21 to 0.39].

For the fine-tuned models, means, jurisdiction, and race are stable [no decision changes when a question is removed; Figure A6]. Their variance results lose significance when any one of four questions is removed, as do employment [1 and 3 questions] and, for the model fine-tuned on 2024 data, ideology [6 questions].

| **Zero-shot test** | **ρ** | **95% bootstrap interval** | **Leave-one-out range** | **Decision changes** |
|---|---|---|---|---|
| H1a: mean answer | 0.71 | 0.34 to 0.90 | 0.67 to 0.76 | none |
| H1b: variance | 0.31 | −0.22 to 0.70 | 0.21 to 0.39 | 3 of 22 |
| H1c-1: city vs. county | 0.88 | 0.67 to 0.95 | 0.86 to 0.91 | none |
| H1c-2: Black vs. White | 0.68 | 0.34 to 0.85 | 0.64 to 0.73 | none |
| H1c-3: conservative vs. liberal | 0.08 | −0.40 to 0.56 | 0.00 to 0.21 | none |
| H1c-4: full-time vs. unemployed | 0.57 | 0.23 to 0.78 | 0.52 to 0.66 | none |

**Table A9.** Zero-shot model, 22 main questions (21 for H1c). Bootstrap interval: percentile interval over 10,000 resamples of questions. Decision changes: number of questions whose removal changes the decision at the registered threshold.

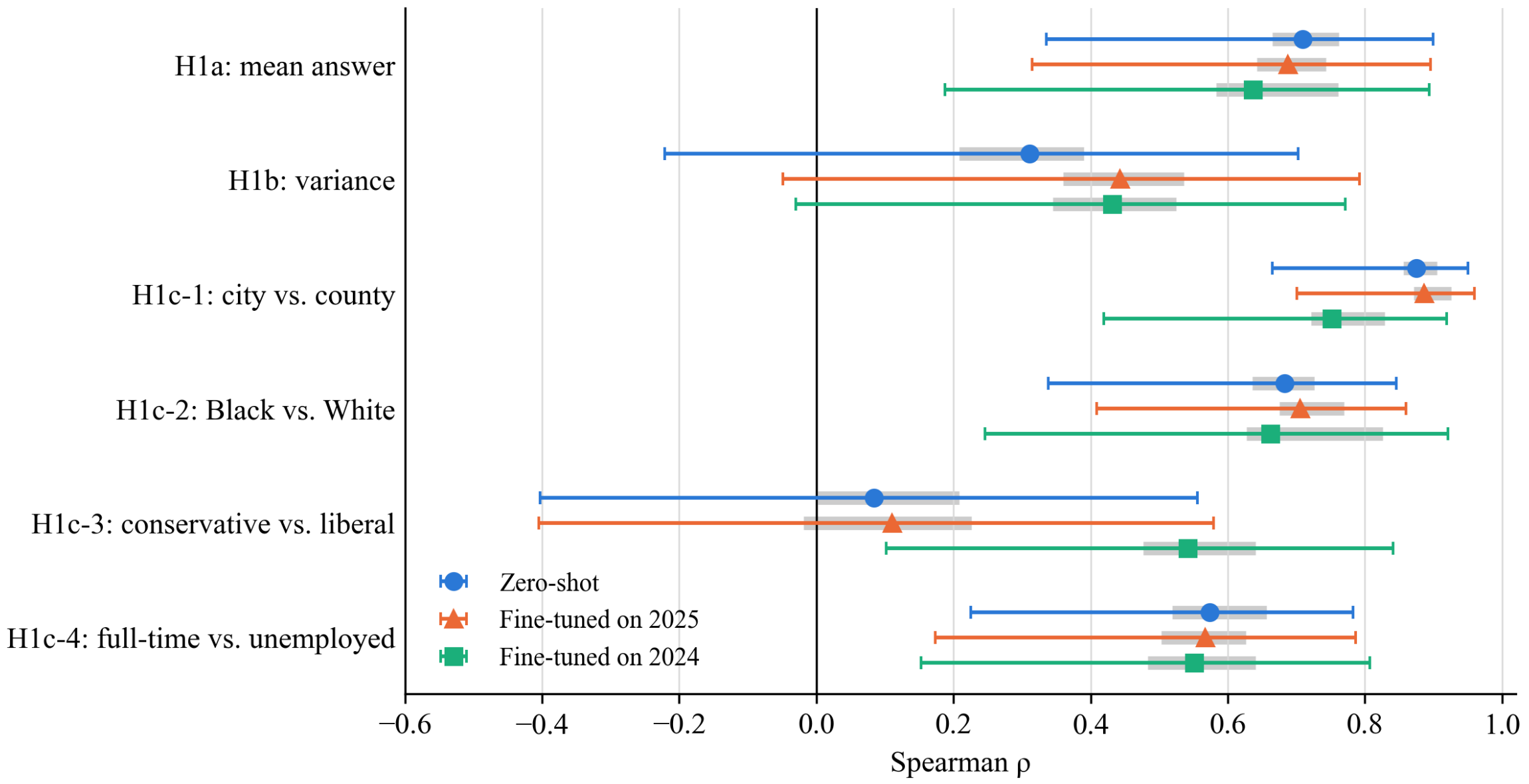

**Figure A6.** Rank correlations for the three models with 95% bootstrap intervals over questions (bars) and the range obtained when each question is removed in turn (grey).

## A13. Split-half reliability of the human rankings

This analysis was not preregistered. We split the 2026 respondents into two random halves 1,000 times, recomputed every observed quantity in each half with the survey weights, and correlated the halves across questions. The Spearman–Brown correction, $2r / (1 + r)$, estimates the reliability of the full sample, and its square root the correlation that a noise-free prediction would reach. Both formulas are approximate for rank correlations.

Table 2 in the main text gives the results for the zero-shot model. The fine-tuned models relate to the ceiling in the same way [correlation with a half-sample, fine-tuned on 2025 and on 2024: means 0.70 and 0.65; variance 0.45 and 0.43; jurisdiction 0.84 and 0.73; race 0.70 and 0.65; employment 0.53 and 0.50; ideology 0.09 and 0.47].

The model fine-tuned on 2024 data correlates with a half-sample on ideology [0.47] more than the other half-sample does [0.35]. This is possible because a prediction carries no sampling noise of its own (Table A10).

| Quantity | Human split-half ρ | ZS: full | ZS: half | FT-25: full | FT-25: half | FT-24: full | FT-24: half |
|---|---|---|---|---|---|---|---|
| Mean answer | 0.98 | 0.71 | 0.71 | 0.69 | 0.70 | 0.64 | 0.65 |
| Variance | 0.92 | 0.31 | 0.30 | 0.44 | 0.45 | 0.43 | 0.43 |
| City vs. county | 0.88 | 0.88 | 0.82 | 0.88 | 0.84 | 0.77 | 0.73 |
| Black vs. White | 0.87 | 0.68 | 0.68 | 0.70 | 0.70 | 0.65 | 0.65 |
| Conservative vs. liberal | 0.35 | 0.08 | 0.03 | 0.10 | 0.09 | 0.56 | 0.47 |
| Full-time vs. unemployed | 0.55 | 0.57 | 0.51 | 0.58 | 0.53 | 0.55 | 0.50 |

**Table A10.** Split-half reliability of the human rankings and, for each model, the correlation with the full 2026 sample and the mean correlation with a half-sample. The same 21 questions are used for every model on group differences, so fine-tuned values can differ slightly from Table 1. ZS: zero-shot; FT-25 and FT-24: fine-tuned on 2025 and 2024 data.

### A14. Baselines from 2025 responses

This analysis was not preregistered. Two baselines predict each new question from 2025 human responses alone: the 2025 question with the most similar text [TF-IDF cosine similarity over title, question, and options; 135 candidate questions], and the average of the 2025 questions in the same module. Table A12 lists the matches. Each baseline supplies a mean, a variance, and a group difference per axis, which are tested like the model's predictions (Table A11).

Neither baseline ranks the means [$\rho = 0.22$ and 0.19 vs. 0.71] or the jurisdiction differences [0.21 and −0.05 vs. 0.88]. The model does not simply repeat the nearest 2025 question [correlation between the two sets of predicted means: 0.37]. Variance is the exception: the nearest-question baseline ranks it better than the model [0.43, $p = 0.022$ vs. 0.31]. The baselines are crude: similarity is lexical and answer scales are not aligned [for example, the four travel-frequency questions are matched to religious attendance; Table A12].

Where an identical 2025 question exists, the model reproduces it. This is the case for one question in the main set: whether windows were open in the past 24 hours was also asked in 2025 and was part of every profile [text similarity 0.91; predicted 0.36, 2025 value 0.36, 2026 value 0.62], and every simulated respondent gave its own 2025 answer. Removing it changes no zero-shot decision [means 0.75; jurisdiction 0.87; race 0.73; employment 0.59, p = 0.003; ideology 0.09; variance 0.21]. The six transportation-insecurity items behave the same way (Appendix A5).

| Quantity (22 main questions) | Zero-shot GPT-4.1 | Nearest 2025 question | Module average (2025) |
|---|---|---|---|
| Mean answer | 0.71 | 0.22 | 0.19 |
| Variance | 0.31 | 0.43 | 0.02 |
| City vs. county | 0.88 | 0.21 | −0.05 |
| Black vs. White | 0.68 | 0.40 | −0.03 |
| Conservative vs. liberal | 0.08 | 0.35 | −0.07 |
| Full-time vs. unemployed | 0.57 | 0.35 | 0.20 |

**Table A11.** Spearman ρ with the observed 2026 quantity on the 22 main questions (21 for group differences), for the zero-shot model and two baselines that use only 2025 human responses.

| New 2026 question | Most similar 2025 question | Similarity |
|---|---|---|
| Windows open in past 24 hours | Open windows in last 24 hours | 0.91 |
| How long windows were open | Open windows in last 24 hours | 0.54 |
| Vacant lots as a problem | Vacant buildings in neighborhood | 0.51 |
| Largest number of windows open at once | Open windows in last 24 hours | 0.46 |
| How wide windows were opened | Open windows in last 24 hours | 0.40 |
| Business satisfaction | Satisfaction with policing | 0.39 |
| Trust in Johns Hopkins University | Trust in local government | 0.34 |
| Ease of finding equivalent home | Job search confidence | 0.33 |
| Business meets family needs | Housing insecurity: payment difficulty | 0.32 |
| Walking frequency past 30 days | Religious attendance | 0.21 |
| Biking frequency past 30 days | Religious attendance | 0.21 |
| Fear of deportation | Financial future: housing | 0.20 |
| Worry about affording to stay | Financial future: housing | 0.19 |
| Own vehicle frequency past 30 days | Religious attendance | 0.16 |
| Public transit frequency past 30 days | Religious attendance | 0.16 |
| Neighbors known by name | Neighbors’ willingness to help | 0.11 |
| Transport trouble for medical appointments | Utility payments | 0.10 |
| Destination if moving | Trust in state government | 0.09 |
| Likelihood of moving in 3 years | Anticipated time to rebuild Key Bridge | 0.09 |
| Preferred residence in 3 years | Gender | 0.07 |
| Business reflects preferences | Neighborhood is welcoming | 0.05 |
| Business influence on community | Neighborhood is welcoming | 0.05 |

**Table A12.** The most similar 2025 question for each of the 22 main questions, by TF-IDF cosine similarity of the question texts, sorted by similarity. The six transportation-insecurity items are in Table A4.

### A15. Procedures for the exploratory and additional analyses

**Full distribution tests.** For each new question, we tested whether simulated and observed responses come from the same distribution, using a two-sample Kolmogorov–Smirnov test on the normalized responses [$\alpha = 0.05$]. This is a stricter criterion than a significant rank correlation across questions, since it requires that the samples are indistinguishable.

**Temporal drift on repeated questions.** Some questions are asked in both the 2025 and 2026 waves. For these repeated questions, we compared the ranking of normalized means and variances in 2025 with the ranking in 2026, among the human responses, to measure how stable rankings are across years in the absence of any model. We kept questions with at least 10% valid responses in both waves and the same set of answer codes in both waves, excluding demographic items [44 questions].

**Six items repeated from 2025.** As preregistered, we recomputed the H1a, H1b, and H1c rank correlations on the full set of 28 questions, which adds the six transportation-insecurity items (Appendix A5).

**Robustness of the variance ranking.** On a bounded scale, variance is partly determined by the mean, and normalized variance also depends on the number of response categories, both of which the simulated and human responses inherit. We therefore also report H1b on the largest fixed-scale subset [the 12 items on a 1–5 scale]. In addition, we report the rank correlation of variance relative to its maximum given the mean, variance / $(m(1 - m))$, where $m$ is the normalized mean.

**Community ratings.** Three questions ask, for each of 11 named neighborhoods, which the respondent does not know anything about, would never consider, or would seriously consider. The unit is the neighborhood. We treated the three items separately. For each item, we computed the weighted share of respondents who selected each neighborhood, ranked the neighborhoods by the predicted and by the observed share, and report the Spearman rank correlation. We also compared the observed shares with the desirability ordering stated in the preregistration. We did not test these correlations against a null hypothesis; with 11 neighborhoods, a rank-correlation test is underpowered. The same analysis was conducted for the seven types of heating in the home.

**Integer-valued questions.** Three new questions ask for integer counts: days worked in the past 14 days, days worked only from home, and household vehicles. For each, we report the predicted weighted mean alongside the observed weighted mean and its 90% confidence interval [weighted bootstrap, 2,000 resamples], and note whether the predicted mean falls within the observed interval. We treat this as a descriptive check on calibration, not a test.

**Controversial, stigmatized, and sensitive issues.** We examined two sensitive questions, fear of deportation and trust in Johns Hopkins University, comparing predicted and observed means and the signed group differences by race and by jurisdiction.

**Attitudes versus behaviors.** We classified the questions as behaviors and experiences [9 of the 22 main questions: window opening, travel modes, and transport trouble for medical appointments] or attitudes and perceptions [13 questions], and repeated the rank correlations within each group. The classification is ours and was not preregistered.

**Signed versus absolute differences.** The preregistered analyses focus on signed coefficients. We also computed the rank correlations on the absolute values of the coefficients, to study whether pre-testing the magnitude of group differences yields different results.

**Analyses added after preregistration.** Three analyses were added after the confirmatory analysis and were not preregistered. First, for every confirmatory test we computed a 95% percentile bootstrap interval for the rank correlation by resampling questions with replacement [10,000 resamples], and recomputed the test with each question removed in turn. Second, to measure sampling noise in the human data, we split the 2026 respondents into two random halves [1,000 splits], recomputed every observed quantity in each half with the survey weights, and computed the Spearman correlation between the halves across questions; the Spearman–Brown correction, $2r / (1 + r)$, gives the reliability of the full sample, and its square root the correlation a noise-free prediction would be expected to reach. Third, we built two baselines from 2025 human responses alone: the 2025 question with the most similar text [TF-IDF cosine similarity], and the average of the 2025 questions in the same module. Results are in Appendix A12–A14.

### A16. Fine-tuned models and pilot

**Fine-tuning.** Three model variants were used: zero-shot GPT-4.1; GPT-4.1 fine-tuned on BAS 2024 respondents; and GPT-4.1 fine-tuned on BAS 2025 respondents. For both years the setup was identical in structure. Respondents were split 85/15 at random. Questions were split by holding out a set of thematic test questions [28 for 2024; 32 for 2025], with all remaining variables used as training context. Each training example was one respondent: the input was the prompt containing demographics and all training-question answers, and the output was that respondent's real answers to the held-out questions. The base model was gpt-4.1-2025-04-14, fine-tuned through the OpenAI API for three epochs with default hyperparameters otherwise, on 85 training and 15 validation respondents drawn from the training set. The resulting models were then used exactly like the base model, with the same prompt structure, to answer the new 2026 questions for all 1,271 simulated respondents.

**Pilot analyses.** To test the viability of the analysis, we ran the process retrospectively on the previous year. Seven LLMs were benchmarked on 32 held-out questions from the 2024 wave, on the rank correlation of means, the ordering of variances, and the direction of group differences across the five axes. GPT-4.1 outperformed the other models on means and group differences and was selected as the primary model [zero-shot: means $\rho = 0.757$, $p < 0.001$; normalized variance $\rho = 0.518$, $p = 0.002$; direction agreement: jurisdiction 30/32, transportation 30/32, employment 28/32, race 24/32, ideology 12/32] (Table A13).

| Axis | Same direction | Questions | Share |
|---|---|---|---|
| Jurisdiction (city vs. county) | 30 | 32 | 94% |
| Transportation (cannot vs. can afford) | 30 | 32 | 94% |
| Employment (full-time vs. unemployed) | 28 | 32 | 88% |
| Race (Black vs. White) | 24 | 32 | 75% |
| Ideology (conservative vs. liberal) | 12 | 32 | 38% |

**Table A13.** Retrospective pilot on the 2024 wave, zero-shot GPT-4.1: number of the 32 held-out questions for which the predicted and the observed group difference have the same sign. In the pilot, the rank correlation was $\rho = 0.757$ for means and $\rho = 0.518$ for normalized variances.